%% file: main.tex
\documentclass[lettersize,journal]{IEEEtran}
\usepackage{amsmath,amsfonts,amssymb,amsthm}
\usepackage{algorithmic}
\usepackage{multirow}
\usepackage{array}
\usepackage[caption=false,font=normalsize,labelfont=sf,textfont=sf]{subfig}
\usepackage{textcomp}
\usepackage{stfloats}
\usepackage{url}
\usepackage{verbatim}
\usepackage{graphicx}
\def\BibTeX{{\rm B\kern-.05em{\sc i\kern-.025em b}\kern-.08em
    T\kern-.1667em\lower.7ex\hbox{E}\kern-.125emX}}
\usepackage{balance}
\usepackage{spconf,epsfig}
\usepackage[usenames,dvipsnames]{xcolor}
\usepackage{soul}
\usepackage{hyperref}
\usepackage{makecell}
\usepackage[switch,columnwise]{lineno}
\usepackage{cite}

\newcommand{\re}[1]{\textcolor{black}{#1}}

\begin{document}



% \maketitle



% \maketitle
\title{Beyond Grid Data: Exploring Graph Neural Networks for Earth Observation}

\author{Shan Zhao, Zhaiyu Chen, Zhitong Xiong~\IEEEmembership{Member,~IEEE}, Yilei Shi~\IEEEmembership{Member,~IEEE}, Sudipan Saha~\IEEEmembership{Senior Member,~IEEE}, Xiao Xiang Zhu~\IEEEmembership{Fellow,~IEEE}
        % <-this % stops a space
\thanks{The work is jointly supported by the German Federal Ministry of Education and Research (BMBF) in the framework of the international future AI lab ``AI4EO -- Artificial Intelligence for Earth Observation: Reasoning, Uncertainties, Ethics and Beyond'' (grant number: 01DD20001), by German Federal Ministry for Economic Affairs and Climate Action in the framework of the ``National Center of Excellence ML4Earth'' (grant number: 50EE2201C), by the Excellence Strategy of the Federal Government and the Länder through the TUM Innovation Network EarthCare, by Munich Center for Machine Learning, and by the TUM Georg Nemetschek Institute Artificial Intelligence for the Built World (GNI) in the framework of ``AI4TWINNING''.}
\thanks{Shan Zhao, Zhaiyu Chen, Zhitong Xiong, and Xiao Xiang Zhu are with Technical University of Munich, Ottobrunn, 85521, Germany. (email: shan.zhao; zhaiyu.chen; zhitong.xiong; xiaoxiang.zhu@tum.de).
\par
Xiao Xiang Zhu is also with Munich Center for Machine Learning, 80333 Munich, Germany.
\par
Yilei Shi is with Technical University of Munich, Munich, 80333, Germany. (email: yilei.shi@tum.de). 
\par
Sudipan Saha is with Yardi School of Artificial Intelligence, Indian Institute of Technology Delhi, New Delhi, 110016, India. (email: sudipan.saha@scai.iitd.ac.in).}}



\maketitle
\begin{abstract}
Earth Observation (EO) data analysis has been significantly revolutionized by deep learning (DL), with applications typically limited to grid-like data structures. Graph Neural Networks (GNNs) emerge as an important innovation, propelling DL into the non-Euclidean domain. Naturally, GNNs can effectively tackle the challenges posed by diverse modalities, multiple sensors, and the heterogeneous nature of EO data.
To introduce GNNs in the related domains, our review begins by offering fundamental knowledge on GNNs. \re{Then, we summarize the generic problems in EO, to which GNNs can offer potential solutions. Following this, we explore a broad spectrum of GNNs' applications to scientific problems in Earth systems, covering areas such as weather and climate analysis, disaster management, air quality monitoring, agriculture, land cover classification, hydrological process modeling, and urban modeling. The rationale behind adopting GNNs in these fields is explained, alongside methodologies for organizing graphs and designing favorable architectures for various tasks. Furthermore, we highlight methodological challenges of implementing GNNs in these domains and possible solutions that could guide future research.} While acknowledging that GNNs are not a universal solution, we conclude the paper by comparing them with other popular architectures like transformers and analyzing their potential synergies. 
\end{abstract}

\begin{IEEEkeywords}
Graph Neural Networks, Earth Observation, Remote Sensing, Machine Learning
\end{IEEEkeywords}

\input{sources/1_introduction}
\input{sources/2_fundementals}
\input{sources/3_solutions}
\input{sources/4_applications}
\input{sources/5_challenges}
\input{sources/6_discussion}

\input{sources/7_conclusions}

\section{Acknowledgements}
\noindent The authors thank Qingsong Xu for sharing relevant literature resources on the application of GNNs in hydrology.
\par

\bibliographystyle{IEEEbib}
\bibliography{main}

\end{document}

%% file: sources/1_introduction.tex
\section{Introduction}
% CNN characteristics
\IEEEPARstart{E}{arth} Observation (EO) data, acquired by various sensors across different locations and times, presents as diverse modalities. Each of these modalities provides unique insights into the status of our planet, facilitating an understanding of the Earth. Deep learning (DL) is particularly appealing in parsing the abundant information in the EO field~\cite{zhu2017, camps2021deep, xiong2022earthnets}. These data-driven models excel in extracting feature representations solely from the data, thereby eliminating the need for manually crafting features based on domain-specific knowledge. When it comes to images, they can be conceptualized as functions in Euclidean space, sampled on a grid where proximity is associated with local connectivity. Convolutional Neural Networks (CNNs) effectively harness the inductive bias of translation invariance and locality by employing convolutions in conjunction with other operations such as downsampling (pooling~\cite{goodfellow2016deep}). The convolution operation allows the extraction of local features shared across the entire image domain, as CNNs apply filters in a ``sliding window'' fashion across the input layer. This approach significantly reduces the number of learnable parameters, facilitating the training of very deep architectures \cite{simonyan2014very}. Moreover, it enables CNNs to learn features independent of the specific region within the input \cite{fukushima1980neocognitron}. These characteristics make CNNs well-suited for handling image datasets in computer vision applications~\cite{he2016deep, ronneberger2015u,chen2017deeplab}. 
% challenges of processing graph using DL
\par However, EO data sometimes deviates from grids or regular formats alike, with an example being Light Detection and Ranging (LiDAR) point clouds, which are discrete points in space without predefined regularity between them. Moreover, EO data, in contrast to traditional vision images, are characterized by specific attributes such as geographical context, spatial-temporal dynamics, adherence to physical processes, and often limited visual saliency. These attributes necessitate a different space to decipher their complexities. Representing such data as graphs offers a compelling alternative, yet effectively handling graph-structured data poses significant challenges. Unlike images, which have a fixed size and spatial characteristics, graphs can exhibit arbitrary structures. They lack a well-defined beginning or end, where connected nodes may not necessarily be spatially adjacent. Besides, nodes in a graph may have a varying number of neighbors, a feature that hinders the straightforward application of two-dimensional (2D) convolution on graphs. Moreover, the traditional DL algorithms often assume that samples are independent of each other. However, this assumption may not hold for graph data, as the samples (nodes) are interconnected through links or edges. This interdependence among nodes adds a layer of complexity to the modeling process~\cite{chakrabarti2006graph, dorogovtsev2000structure}.
\par
% GNNs
In response to the challenges posed by graph data, a novel class of methods has emerged, known as Graph Neural Networks (GNNs)~\cite{scarselli2008graph}. The essence of GNNs lies in the interactive aggregation and processing of the information across nodes and their connectivities (edges). These innovative approaches involve the development of new definitions and the generalization of existing operations from Euclidean space to non-Euclidean space. For instance, traditional 2D convolution is replaced with techniques better suited for graph structures~\cite{kipf2016semi, defferrard2016convolutional}. The advent of GNNs represents a paradigm shift in the field, enabling machine learning (ML) and DL models to effectively capture and leverage the complex relationships within graph data. 
% GNNs for EO applications
\par
\re{The unique properties of EO data, coupled with the strengths of GNNs, have catalyzed the rapid adoption of GNNs in the EO field. The earliest and most straightforward use of GNNs in this field was through semi-supervised learning for fundamental remote sensing (RS) tasks such as image classification~\cite{mou2020nonlocal, ren2021semi, li2021manifold}, segmentation~\cite{shi2020building}, and change detection~\cite{saha2020semisupervised, tang2021unsupervised}. Their ability to extract long-range contextual information and efficiency in organizing superpixels from grid-based inputs has made GNNs particularly prevalent in hyperspectral image analysis~\cite{liu2020semisupervised, yu2023uncertainty}. Over time, GNNs have also been applied to specialized EO data types with inherent graph structures, such as those found in scattered monitoring stations, river networks, and road layouts. This has expanded their use in air quality monitoring~\cite{xiao2022dual, qi2019hybrid}, flood detection~\cite{kazadi2022flood, li2020hybrid}, and urban planning~\cite{chu2019neural, he2020sat2graph}. Additionally, GNNs have shown promise in addressing global weather forecasting~\cite{keisler2022forecasting, lam2022graphcast}, leading the way toward replacing traditional physical models with purely data-driven approaches. Furthermore, GNNs are adapted for dynamic complex systems. They are being explored for tasks such as natural resources exploitation~\cite{guan2022recognizing}, crop yield forecasting~\cite{qiao2023kstage}, seismic activity prediction~\cite{zhang2022spatiotemporal}, and other areas where understanding complex, interconnected processes is essential.}
\par
% Difference to previous work
The target of this survey is to provide an overview of GNN applications in the EO field. While many reviews on GNNs exist within the computer vision community~\cite{GNNBook-ch8-gunnemann, han2021dynamic, zhou2022graph, bo2023survey, zhou2020graph}, they often fall short of providing relevant guidance to geoscientific communities due to the significant gap between the two domains. Besides, although several reviews on the application of GNNs in the EO domain are available~\cite{khlifi2023graph}, they predominantly concentrate on traditional tasks such as RS image classification, segmentation, and change detection. However, real-world geoscientific problems are far more complex than these simplistic scenarios. Thus, there is a pressing need for a review that emphasizes geoscientific challenges and extends its focus beyond RS imagery to encompass diverse data sources such as monitoring stations and climate projections. More importantly, the challenges of implementing GNNs in EO have not been well summarized. Therefore, we recap fundamental generic problems from both geoscientific and methodological views, along with the promising research directions motivated by addressing these obstacles. The contributions of our work include
\begin{itemize}
    \item We provide a comprehensive review of GNNs on EO, with applications in atmospheric science, biogeosciences, climate, hydrological sciences, hazards, and seismology communities.
    \item We summarize the key generic problems shared across the above domains and demonstrate how GNNs can effectively address them.
    \item We analyze the major impediments preventing wider adoption of GNN-based methods in the EO domain, and offer potential solutions for overcoming these challenges.
    \item We illustrate mind maps for conceptualizing and designing GNNs in the EO domain, and analyze the synergies between GNNs with other prominent DL models to encourage innovative solutions for real-world deployment.
\end{itemize}
\par
The paper is organized as follows. \re{Chapter~\ref{sec:fundamental knowledge} presents fundamental background information on EO data types and GNNs. This chapter examines the limitations of traditional DL techniques in processing EO data and illustrates how GNNs can effectively address these challenges. In Chapter~\ref{sec:problems}, we provide an in-depth discussion on the common challenges associated with EO data shared across different EO-related use cases, and the advantage of GNNs in overcoming them. In Chapter~\ref{sec:applications}, we delve into the case studies exemplifying the superiority of GNNs in addressing each EO application, including the methodologies for organizing EO data as graphs, designing appropriate architectures, and experimental results that show the real-world implications of GNNs. In Chapter~\ref{sec:challenges}, we outline the major barriers to implementing GNNs in the EO domain, followed by potential solutions to narrow the gap between technical and domain science communities. Chapter~\ref{sec:discussion} compares GNNs with other prevalent architectures, such as CNNs and Transformers, exploring the opportunities for hybrid models. The paper concludes with future research directions of developing GNNs in the context of EO.} In this contribution, we aim to assist domain experts in conceptualizing graph structures and designing suitable model architectures given EO data. Synergically, the methodological advance of GNN tools and domain-specific analysis demands opens up new areas of understanding our dynamic Earth.

%% file: sources/2_fundementals.tex
\section{Fundamental knowledge on EO and GNNs}
\label{sec:fundamental knowledge}
\subsection{Earth Observation Data}
\begin{figure}[!t]
    \centering
    \includegraphics[width=1\columnwidth,trim={0cm 4.8cm 9.6cm 0cm},clip]{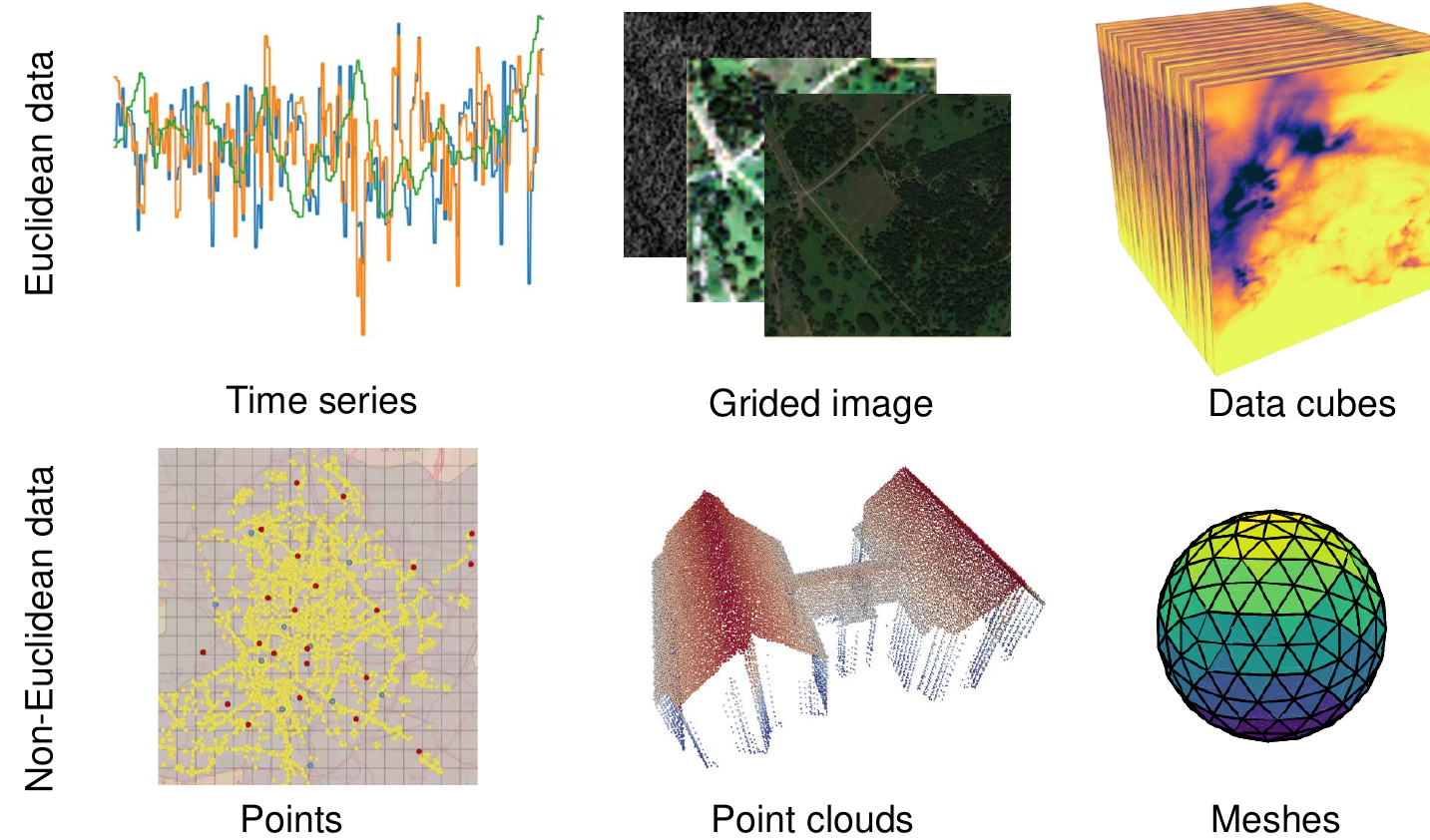}
    \caption{Multiple types of data are collected by EO sensors. For example, in Euclidean space, data can be organized as times-series (e.g., teleconnection indices), grided images, and datacubes (e.g., hyperspectral imagery, spatial-temporal images). The non-euclidean data includes points (e.g., monitoring stations, road sensors, meteorological stations), point clouds (e.g., LiDAR), mesh representations, etc.}
    \label{fig:eu_noneu_data}
\end{figure}

The advent of sensing techniques has broadened the spectrum of data sources accessible for EO. These sources encompass a diverse array of formats, as Fig.\ref{fig:eu_noneu_data} demonstrates.
\par
% Euclidean data.
DL has significantly impacted the field of EO, where most of the data is organized in a Cartesian coordinate system and where network architectures are designed to model the invariances of these structures~\cite{bronstein2017geometric}. Euclidean data is data that follows the principles of Euclidean geometry and can be sensibly modeled in n-dimensional linear space, ranging from sequential time series data, satellite imagery, and grided measurements, to hyperspectral imagery. \re{While the locality and invariance of convolutional kernels significantly reduce the number of parameters required for training, they often fall short in capturing the complex spatial-temporal characteristics inherent in EO data. Specifically, each data type presents unique challenges when processed with traditional DL methods. Below, we discuss these limitations and how GNNs offer promising solutions.
\begin{itemize}
    \item Sequential Time Series:
    Temporal features are crucial for analyzing the dynamics of Earth events. Traditional models like one-dimensional (1D) CNNs and Recurrent Neural Networks (RNNs) are commonly used to process sequential data, but they often neglect spatial patterns and constrain the input to a strict sequential order. In contrast, GNNs allow for capturing spatial and temporal patterns simultaneously. By representing each sequence as a node in a graph, GNNs can integrate multiple sequences and facilitate the exploration of graph-level attributes, thereby simplifying the identification of common or abnormal signal among sequences. Additionally, when each time step is represented as a node, the flexible design of connectivity enables the arbitrary interactions among time steps.
    \item Remote Sensing Images:
    CNNs are well-suited for processing regular inputs like RS images; however, their convolutional kernels are limited to local receptive fields, which restricts the model's ability to capture broader contextual information. GNNs address this limitation by incorporating spatial relationships into the graph structure or by facilitating information exchange between distant neighbors. Furthermore, traditional CNN layers and downsampling layers often produce overly smoothed outputs, which is particularly problematic in tasks requiring precise boundaries such as urban mapping or disaster monitoring. GNNs, by adaptively learning the significance of different nodes and edges, provide a more detailed representation of spatial features.
    \item Global Gridded Inputs:
    In weather and climate analysis, EO data are often mosaicked from dispersed stations and then resampled into a uniform grid covering the entire globe. While these grids simplify the processing with traditional models, it introduces training biases towards polar areas. GNNs can mitigate this issue by mapping the grids back to a spherical representation, thereby aligning the data more closely with its physical properties.
    \item Data Cubes:
    The joint analysis of multiple Earth snapshots, or data cubes, is valuable for monitoring changes over time. Although video-analysis techniques like ConvLSTM~\cite{shi2015convolutional} can be applied, they are still constrained by the limitations inherent in convolutional operations. GNNs, especially their spatiotemporal variants, offer a better understanding of the evolution of Earth phenomena.
\end{itemize}}

\par
% Non-Euclidean data. Details on Point cloud and mesh.
On the contrary, a substantial segment of geospatial research investigates EO data characterized by an inherent non-Euclidean spatial structure. \re{This includes diverse data types such as LiDAR point clouds, Global Navigation Satellite System (GNSS) points, ship trajectories~\cite{feng2022stgcnn, yang2022multi}, geo-tagged social media data~\cite{rode2022true, zhu2022}, and sensor networks. Traditional DL models struggle with directly handle such data. They often require specialized preprocessing steps like voxelization. However, GNNs offer a robust solution for handling such irregularity.
\begin{itemize}
    \item Point Clouds: 
    Point clouds are particularly prevalent in urban environment modeling. A point cloud is a collection of data points in a three-dimensional space, typically scattered on surfaces of objects and scenes. Each point in the point cloud is identified by its position $(x, y, z)$ in three-dimensional (3D) space, and may also include auxiliary information such as intensity or color. Point clouds are commonly obtained through 3D scanning technologies like LiDAR, photogrammetry, or synthetic-aperture radar tomography (TomoSAR). The representation of point clouds as graphs is a natural fit due to the inherent spatial relationships between points. In this representation, each point serves as a node, and the spatial connections between points are captured by edges in the graph. 
    \item Vector Data: 
    Vector data, such as networks of rivers or transportation systems, has inherent graph topology. GNNs can incorporate domain-specific knowledge into their architecture. Therefore, the alignment between the data structure and the model architecture allows for the enhanced physical feasibility of the models.
    \item 3D Mesh Data:
    Similarly, 3D mesh data, depicting surfaces through interconnected vertices, edges, and faces, provides a structured representation of 3D objects. Mesh can also be represented as a graph, with vertices as nodes and edges connecting them.
\end{itemize}
In summary, the diverse data types provide complementary perspectives on our Earth, and GNNs are well-suited to address the irregularity, physical consistency, dynamics, and complexity in EO data.}
\subsection{Graph Neural Networks}
\subsubsection{Graphs}
% Definition of graph
Formally, the two components of a graph $\mathcal{G}=(\mathcal{V}, \mathcal{E})$ are vertices (nodes) $\mathcal{V}$ and edges $\mathcal{E}$. There are various types of $\mathcal{E}$; for example, they can be weighted or unweighted, directed or undirected, depending on the given application~\cite{banerjee2017graph}.
\par 
% Graph representation: matrix, list
A prevalent approach to denoting a graph structure is using an adjacency matrix $\mathcal{A}$. In the context of a graph with $N$ nodes, the corresponding adjacency matrix $\mathcal{A}$ takes the form of an $N$ $\times$ $N$ matrix, where each row and column represents an individual node, and the intersected value is the connectivity between these two nodes. To mitigate computational complexity, a sparsity matrix is often employed, retaining only the indices and values of the non-zero edges. Another approach to represent the graph is the adjacency list, where the neighbors of each node or all edges between node pairs are stored.
\subsubsection{GNNs}
% Node/Edge embedding; message passing; pooling.
GNNs extend DL to graph-structured data. As part of feature engineering, the first step often involves projecting nodes and edges into a feature space, a process known as node embedding or edge embedding. These embeddings consist of vector representations of node or edge properties within a graph, serving as ingredients for subsequent GNN training.
\par
\noindent\textbf{Message passing:}
%  Message passing (Neighborhood aggregation): spatial and spectral graph convolution.
The key idea in GNNs is the differentiable message passing. Consider a graph $\mathcal{G}=(\mathcal{V}, \mathcal{E})$, where $e_{ij}$ is the edge weight between node $v_i$ and node $v_j$. Every node $v_i$ has feature vector $h_i \in \mathbb{R}^d$. Let $h_i^{l-1}$ be the hidden representation of node $v_i$ at the $l-1$ layer. For each node, two operations are conducted:
\par
The first is to gather messages from all neighbors $\mathcal{N}$,
\begin{equation}
    m_i^{(l)} = \sum_{v_j \in \mathcal{N}(v_i)}f_M(h_i^{(l-1)},h_j^{(l-1)},e_{ij}).
\end{equation}
\par
\begin{figure}[!t]
    \centering
    \includegraphics[width=1\columnwidth,trim={0cm 0.3cm 1.2cm 0cm},clip]{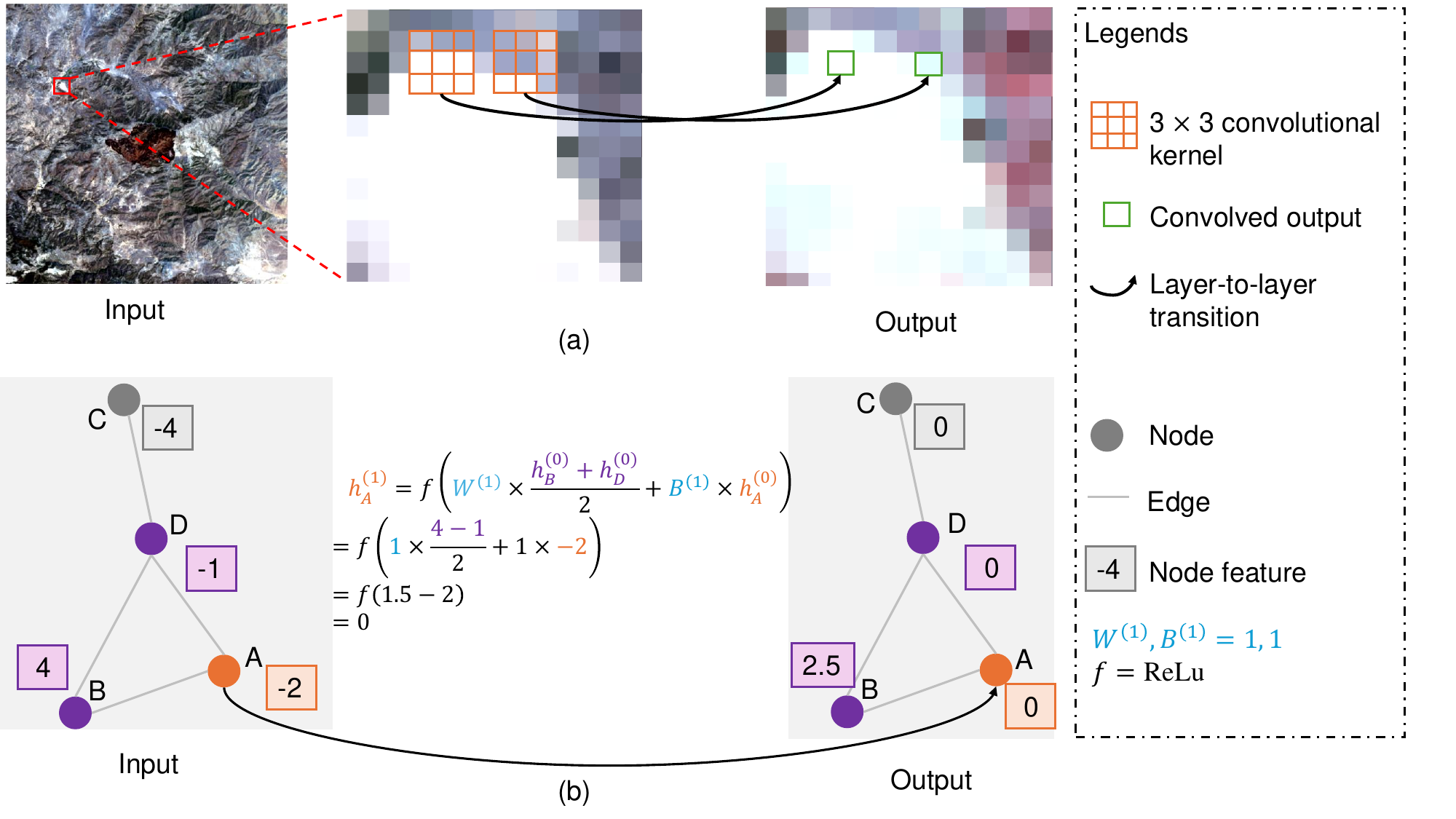}
    \caption{Comparison between a CNN layer and a GNN layer. (a) CNNs update inputs through a shift-invariant weighted sum within a regular window. (b) GNNs update inputs by aggregating the information from their neighbors. For clarity, we use a single number to represent the node feature. The example demonstrates how node features are updated using a simple GCN layer. }
    \label{fig:cnn_vs_gnn:convolution}
\end{figure}

The second is to update the hidden representation of layer $l$,
\begin{equation}
    h_i^{(l)} = f_U(h_i^{(l-1)},m_i^{(l)}),
\end{equation}
% spatial gnn
where $f_M,f_U$ are any differentiable functions, e.g. Neural Networks (NNs). Message passing is explicitly implemented via spatial graph convolution, where the operation involves direct and localized node-to-node communication. For example, Graph Attention Networks (GATs)~\cite{velivckovic2017graph}, GraphSAGE~\cite{hamilton2017inductive}, and Message Passing Neural Networks~\cite{gilmer2017neural} leverage the intrinsic relationships and features within the graph's topology for enhanced data processing and analysis. Figure~\ref{fig:cnn_vs_gnn:convolution} compares the evolution of hidden states through a single-layer CNN and a single-layer GNN.

\par
% spectral gnn
Another effective implementation is the spectral GNNs~\cite{bruna2013spectral}, which deal with the properties and analysis of graphs through the lens of eigenvalues and eigenvectors. In spectral graph convolution, the convolution operation is defined in the spectral domain by multiplying the Fourier transform of the graph signals (node features) with a filter. Bianchi \textit{et al.}~\cite{bianchi2021graph} employ AutoRegressive Moving Average (ARMA) filters in the spectral domain for graph convolution, allowing for more flexible and powerful spectral filtering. In comparison to spatial graph convolution, spectral graph convolution excels at capturing the global structure of the graph. To reduce the cost associated with eigenvalue and eigenvector computation, Chebyshev polynomials have been employed to approximate the spectral filters of the graph Laplacian~\cite{defferrard2016convolutional}. Kipf and Welling~\cite{kipf2016semi} propose a simplification of the convolutional operation via a localized first-order approximation, known as Graph Convolutional Networks (GCNs).

\par
\begin{figure}[!t]
    \centering
    \includegraphics[width=1\columnwidth,trim={0cm 0cm 2.5cm 0cm},clip]{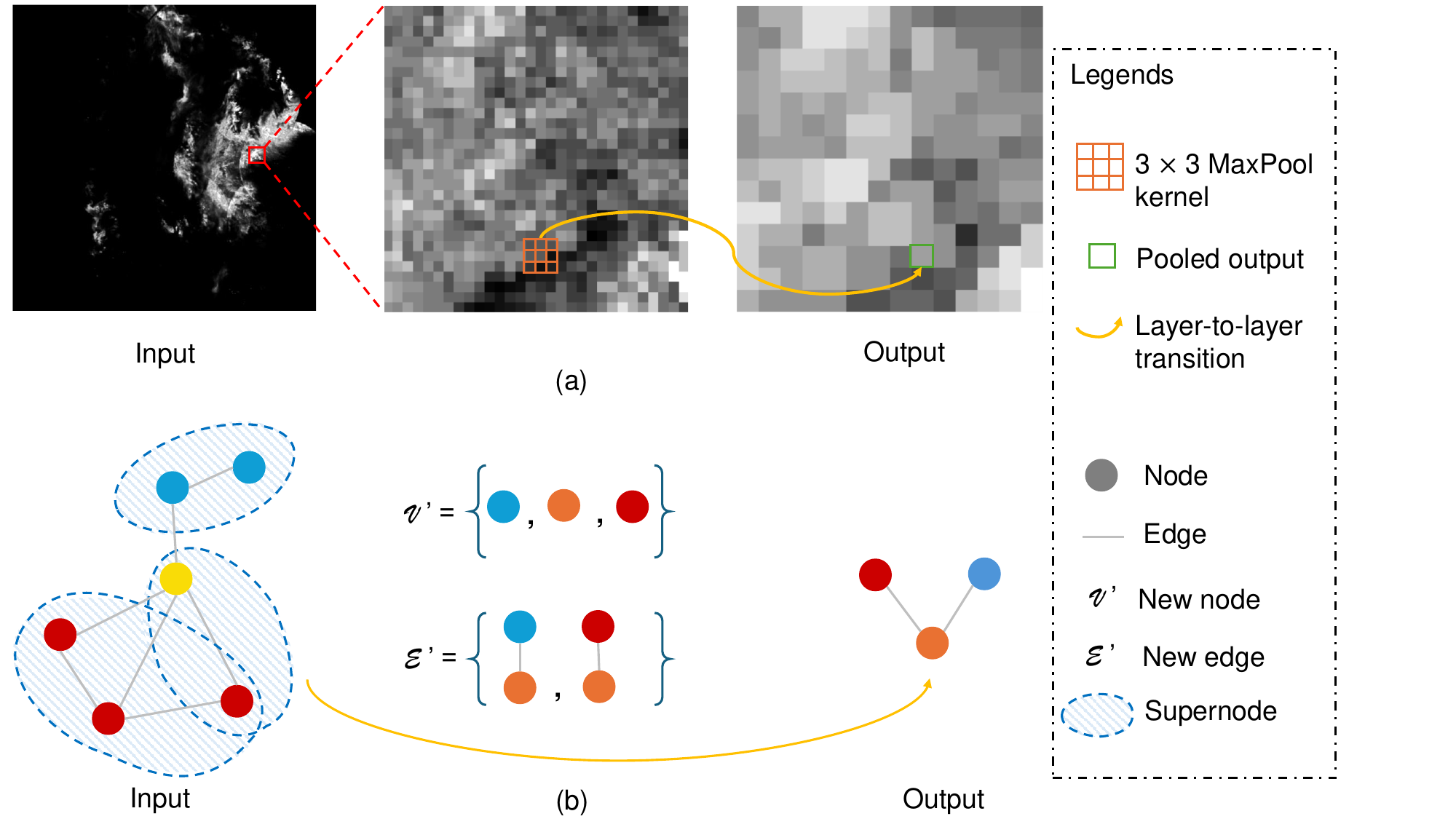}
    \caption{The comparison between a traditional pooling layer and a graph pooling layer. (a) Max pooling with 2 strides on a sample Radar map. The pooling on grided data only changes the value/number of elements. (b) The graph pooling requires the update of graph topology due to the change of node numbers.}
    \label{fig:cnn_vs_gnn:pooling}
\end{figure}

\noindent\textbf{Pooling:}
% pooling
Pooling is an important operation in CNNs to reduce dimensionality and improve computation efficiency; moreover, it prompts translation and feature invariance. In GNNs, there are similar layers to help reduce the graph size and improve graph generalizability. To get a coarsened graph, the pooling operation firstly selects subsets of nodes from the whole graph, then aggregates each subset to an output node, and finally connects reduced nodes with edges~\cite{grattarola2022understanding}. A graph pooling layer can be represented as
\begin{equation}
    \mathcal{V}', \mathcal{E}' \longleftarrow \textbf{Pool}(\mathcal{V}, \mathcal{E}).
\end{equation}
% global pooling, hierarchical pooling
Global pooling uses a set of readout functions such as global add, mean, or max pooling to condense a graph to a single node. Hierarchical pooling, on the contrary, provides a multi-resolution representation of the graph. This also allows deeper GNN models. Figure~\ref{fig:cnn_vs_gnn:pooling} compares the pooling operation on grided data and on graph-structured data.
\par

\begin{figure*}[!t]
    \centering
    \includegraphics[width=2\columnwidth,trim={0cm 0cm 0cm 0cm},clip]{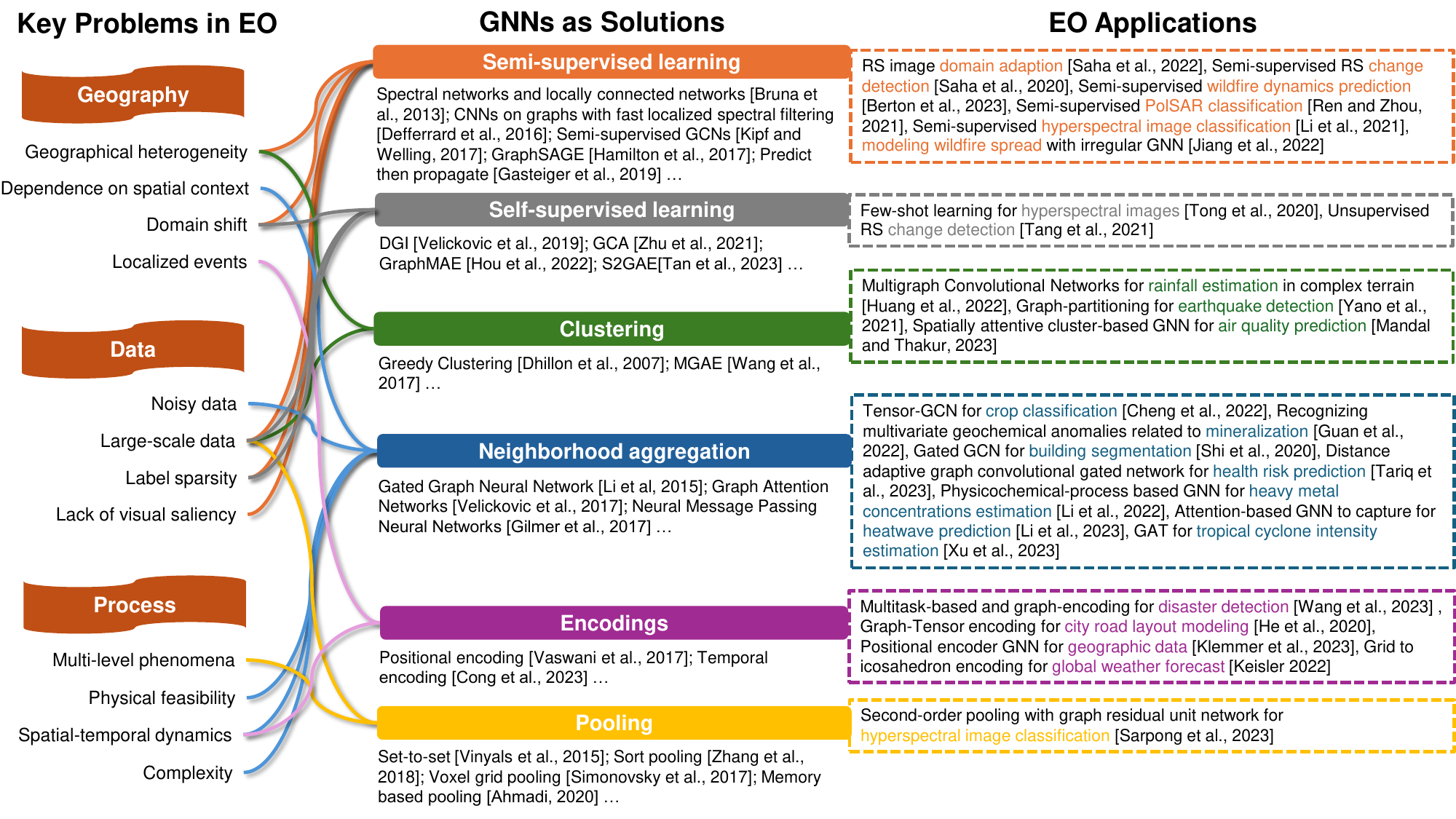}
    \caption{\re{Variants of GNNs: Advanced GNNs, their unique advantages to solve EO Challenges (Chapter~\ref{sec:problems}), and examples of successful applications in the EO Field (Chapter~\ref{sec:applications}).}}
    \label{fig:gnn_variants}
\end{figure*}
% \par
% Table: overview of applications
\re{The past few years have witnessed rapid advancements in vision GNNs, whose application to EO tasks effectively breaks through some bottlenecks of traditional DL methods. Table~\ref{tab:GNNsvsTraditional} presents a comparison of the limitations in traditional DL methods and highlights the advantages of GNNs as emerging solutions for handling EO data.}
%table inspried by~\cite{reichstein2019deep}
\begin{table*}[!t]
\caption{\re{Conventional DL and GNN-based Approaches to EO Tasks. GNN-based Methods Emerge as Potential Solutions to Address the Unique Challenges Posed by EO Data, Surpassing the Limitations of Traditional DL Techniques.}}
\label{tab:GNNsvsTraditional}
    \centering
    \begin{tabular}{p{1.8cm} p{2.5cm} p{3.5cm} p{3cm} p{4.2cm}}
    \hline
       \textbf{Task} & \textbf{EO task} & \textbf{Conventional DL approaches} & \textbf{Limitation of conventional approaches} &\textbf{Advantage of GNN-based emergent or potential approaches} \\
       \hline
       \multirow{3}{*}{Classification} & LCLU classification using hyperspectral images & 1D, 2D, 3D CNNs, RNNs  & Limited spatial context & Local and global contextual information extraction~\cite{mou2020nonlocal}, multi-scale feature extraction~\cite{jia2022graph}\\ \cline{2-5}
       ~ & Out-of-domain multi-spectral image classification & Generative modeling, adversarial training  & Separate model training for multiple targets & Message Passing and Label Propagation~\cite{saha2022multitarget, zhao2022graph} for better generalizability \\ \cline{2-5}
       ~ & Crop classification using SAR & CNNs & No global information, time-varying features & Global optimal neighborhoods capturing~\cite{cheng2022novel}, robust to noise by combining neighborhood information\\ \hline
       \multirow{3}{*}{Segmentation} & Building footprint generation & CNNs (FCN, U-Net, SegNet...) & Limited and local receptive field, blurred boundaries & Model local details from neighbors~\cite{shi2020building}; preservation of spatial fine-granularity~\cite{li2020building}\\ \cline{2-5}
      ~ & Point cloud semantic segmentation; building segmentation & Point-based neural networks & High computational cost & Processing unstructured data \\ \cline{2-5}
      ~ & Digital elevation/terrain map extraction & Multi-view-based, voxel-based, and point-based neural networks & Unable to handle occlusion situations, high computational cost & Improved efficiency in processing unstructured data, dealing with sparse ground points and complex terrains with steep slopes~\cite{zhang2020extraction}\\ \cline{2-5}
       ~ & Disaster mapping & CNNs, RNNs & Poor generalization & Node density adjusting for fast damage mapping~\cite{jiang2022modeling}, addressing environmental heterogeneity~\cite{zeng2022graph}, generalizing to post-disaster images\\ \hline
       Denoising & SAR image denoising & Statistical based method & Sacrifice of spatial resolution, partially fulfilled assumptions & Superpixel-based nodes to against speckling effects~\cite{wang2021dynamic}\\ \hline 
      \multirow{2}{*}{Change detection} & Semi-/un-supervised change detection for VHR and HR & Siamese CNNs, RNNs & Susceptible to variations in sensing conditions, fixed and limited receptive field,  often lack contextual information & Capturing long-range contextual patterns~\cite{su2022nonlocal}, non-locality~\cite{mou2020nonlocal}\\ \cline{2-5} 
      ~ & Disaster change detection & Difference images based neural networks & Local spatial neighborhood, lopsided prediction results in complex environments & Capturing similar conditions in long-ranges~\cite{zeng2022graph}; modeling of the hierarchy of damage levels~\cite{wang2023hierarchical}\\ \hline 
      Data fusion & LiDAR and hyperspectral data fusion & DNNs (deep belief networks, deep Boltzmann machines, stacked autoencoders), CNNs & Huge
    number of parameters  & Graph-based feature fusion \cite{ghamisi2016hyperspectral}\\ \hline 
      Video prediction & Weather prediction & Transformers, diffusion models & Heavy computational cost, imbalanced training due to projection distortion  & Capturing of dynamic range interactions~\cite{lam2022graphcast}; addressing polar singularities~\cite{weyn2020improving}\\ \hline 
      \multirow{6}{*}{Regression} & Seismic event prediction, detection, and source characterization & 1D CNNs & Less robust, no spatial information & Spatial-temporal dynamics capturing~\cite{zhang2022spatiotemporal}, leveraging information from multiple stations for improved robustness against noisy signals.\\ \cline{2-5} 
       ~ & Air quality forecasting & CNNs, RNNs, and the combination of both (Convolutional LSTM) & Difficult to handle missing data and non i.i.d data & Handling irregular stations~\cite{tariq2023distance}, capturing spatial relationship~\cite{xiao2022dual, iskandaryan2023graph}; handling spatial-temporal dynamics~\cite{qi2019hybrid, ge2021multi, wang2020pm2}, new area prediction~\cite{tariq2023distance} \\ \cline{2-5}
       ~ & Heavy metal concentrations estimation & Back propagation NN, radial basis function NN, generalized regression NN, MLP & Independence assumption on inputs & Graphs to model chemical processes and correlation betweeen factors~\cite{li2022field} \\ \cline{2-5}
       ~ & Crop yield prediction & CNNs, RNNs (GRUs, LSTM) & No spatial correlation, sequential order of inputs & Spatial-temporal dynamics modeling~\cite{qiao2023kstage, fan2022gnn}\\ \cline{2-5}
       ~ & Heatwave prediction & CNNs & Unable to handle spatial discontinuity of data & Capturing interplays among meteorological features, handing irregular station distribution~\cite{li2023regional} \\ \cline{2-5}
       ~ & Cyclone intensity estimation & CNNs, RNNs, generative adversarial networks & Coarse-grained image recognition, only final wind speed produced & Prior knowledge distillation through probability transition graph, fine-grained feature extraction~\cite{xu2023tfg} \\ \hline
       Anomaly detection & Mineralization exploitation & Autoencoders & Cannot extract deeper features from complex geological systems, cannot represent the spatial structure of geochemical patterns & Modeling of the spatial correlations~\cite{guan2022recognizing} \\
    \hline
    \end{tabular}
\end{table*}

%% file: sources/3_solutions.tex
\section{Key Problems in EO and GNNs as Solutions}
\label{sec:problems}
EO data exhibits unique characteristics such as spatial autocorrelation, complexity, irregularity, and multi-modality. Additionally, it requires specialized expert knowledge for effective processing and analysis due to the complex physical processes involved. \re{This further complicates the acquisition of labeled data for efficient supervised learning. }In this section, we identify five important generic challenges common in the EO field. \re{While traditional DL methods often struggle with these challenges, GNNs offer promising solutions due to their ability to model complex relationships, provide contextual information, and process non-Euclidean data}. In general, the decision to model the EO data using GNNs arises from two aspects. On the one hand, GNNs overcome the CNNs' limitation in processing the non-Euclidean data. On the other hand, gridded data can be conceptualized as graphs to address the unique EO-related challenges.
\par
\re{Figure~\ref{fig:gnn_variants} provides an overview of the rationale behind selecting specific GNN architectures to tackle various EO challenges, supported by successful application cases for reference.}

% Unique EO challenges that GNN is able to solve
\subsection{Geographical Dependence}
Unlike common vision data features that exhibit positional invariance, geographical features are often associated with specific geographic locations. While most convolutional kernels are shift-invariant, they may struggle to capture location-dependent variations. GNNs address these variations across longitudinal and latitudinal dimensions by 1) considering spatial autocorrelation into adjacency matrix design~\cite{shi2022gnn}, 2) using positional-agnostic nodes or augment node features with location information~\cite{bloemheuvel2201multivariate}, and 3) capturing long-distance contextual-aware information~\cite{su2022nonlocal}. 
\par
The distribution of EO data often exhibits spatial autocorrelation, where similar values tend to cluster together. To quantify this similarity between nodes, a distance-based adjacency matrix can be employed~\cite{guan2022recognizing, shi2022gnn,pei2023application}, promoting information aggregation among similar objects. However, spatially adjacent values may not necessarily be similar; hence, the utilization of heterogeneous graphs and multiple graphs is encouraged~\cite{huang2022multigraph, maurya2021improving}. These approaches are well-suited for handling geometrically and topologically complex domains to satisfy spatial and locality requirements.
\par
% position-variant
Furthermore, many Earth phenomena show strong affiliations to geographical positions~\cite{zhu2024foundations}. For instance, a temperature anomaly in the North Pacific should be distinguished from one in the Arctic. While most convolutional kernels are shift-invariant, they may overlook location-dependent variations. GNNs excel in mapping observations from different locations. This is achieved by projecting different regions to distinct nodes or incorporating location information as node or edge attributes~\cite{bilal2022early, bloemheuvel2201multivariate, shi2022gnn}. Weyn \textit{et al.}~\cite{weyn2020improving} train separate weights for faces on the equator and at polars to capture the markedly divergent evolution of weather patterns across different geographical faces. Additionally, to harmonize atmospheric motions in the Arctic with those in the Southern Pole, an ingenious flipping is conducted preceding the convolution operation. These approaches enhance the modeling of location-specific characteristics of EO data.
\par
% contextual-information
Geographical phenomena often converge over larger spatial and temporal coverage. Consequently, it is essential to consider the broader contextual information. GNNs offer a solution to incorporate long-distance information flow and nonlocal features~\cite{su2022nonlocal}. Positional Encoder GNN (PE-GNN)~\cite{klemmer2023positional} explicitly incorporates spatial context embedding through learning sinusoidal transformations of spatial coordinates at different frequencies. In weather forecasting~\cite{lam2022graphcast}, message passing among icosahedral meshes of increasing resolution facilitates spatial interaction across various ranges. More importantly, GNN-based simulators capture spatial interactions over significantly longer distances than traditional methods, thereby supporting coarser native timesteps. To overcome the limited receptive field of CNN kernels, Li \textit{et al.}~\cite{li2023dci} introduce a global maximum node connection algorithm for distant context-aware feature extraction and aggregation. Such long-range extraction capability ensures accurate and comprehensive capture of the spatial and temporal spread of events.

\subsection{Complex Network Analysis}
The occurrence of certain phenomena on Earth is often influenced by multiple variables that exhibit complex and interconnected relationships. Complex network analysis offers an effective approach to unravel these complex interaction patterns. Commonly used graph analysis based on correlation coefficient is prone to pseudolinks~\cite{boers2019complex}. 
\par
Though there are promising tendencies of analyzing causal graphs~\cite{runge2015identifying, runge2015quantifying}, GNN-based graphs lead to another possible interpretation. For instance, graph measures of the adjacency matrix manifest the emergence mechanism of El Ni$\Tilde{\text{n}}$o events~\cite{cachay2021world}. The high eigenvector centrality near the Oceanic Ni$\Tilde{\text{n}}$o Index (ONI) region for one lead month serves as a triggering signal for the El Ni$\Tilde{\text{n}}$o events, while a more widespread distribution of centrality over longer horizons explains how the El Ni$\Tilde{\text{n}}$o phenomenon aggregates from drivers across broader geographical ranges.

\subsection{Irregular, \re{Heterogeneous, and Noisy} Data}
Given the wealth of available data sources, they often vary significantly in terms of coordinates, resolution, temporal and spatial coverage, \re{quality,} and modalities. While these diverse data sources offer the potential for richer perspectives on phenomena, integrating them into an end-to-end DL framework is inherently demanding. GNNs offer a solution to organize irregular data. They can manage unstructured data, deal with variables of various types and sources~\cite{zeng2022graph}, accommodate variables with varying resolutions in space, address data distortions caused by transforming from spherical coordinates to a planar coordinate system~\cite{shi2022gnn}, \re{and contribute to robust prediction by handing noisy inputs}. 
\par
To integrate variables from heterogeneous sources, Zhang \textit{et al.}~\cite{zhang2023deep} differentiate nodes by their types. Up to 78 atmospheric variables with auxiliary variables are attached as initial node features. In~\cite{kazadi2022flood}, node feature types transcend scalar data and extend to vectors. The vector data, such as water flow velocities, are processed by Geometric Vector Perceptrons in order to preserve their geometric properties. 
\par
GNNs also offer solutions to handle varying resolutions in space, often achieved by adopting hierarchical graph structures. Shi \textit{et al.}~\cite{shi2022gnn} store a graph hierarchy by graph coarsen, and generate adaptive resolution outputs for future simulation runs according to regions sensitivities. MultiScaleGNN~\cite{lino2022towards} partitions the whole space into grid cells at different levels, allocating more computational resources to regions where the physics is complicated to resolve, and fewer resources elsewhere. GraphCast~\cite{lam2022graphcast} supports multi-scale mesh representation using GNN-based processors~\cite{keisler2022forecasting}. These graph structures effectively capture both local and global scale atmospheric processes.
\par
Noise is an inherent challenge in EO data analysis, arising from various stages of data acquisition and influenced by factors such as sensing conditions and data transmission errors. GNNs reduce noise by integrating information from multiple sources. GNNs' aggregation mechanism, which functions similarly to ``majority voting" or ``ensemble prediction", leads to more robust output. For instance, synthetic-aperture radar (SAR) images are often affected by speckle noise, but by organizing superpixels as nodes, GNNs can provide predictions based on a more resilient representation of geographical units~\cite{wang2021dynamic}. Similarly, point clouds can be organized into superpoints, allowing for efficient processing with a compact representation~\cite{landrieu2018large}. In seismic characterization, where traditional single-station inverse problems are vulnerable to signal noise, GNNs offer a solution by analyzing multiple time series and regressing graph-level attributes. At a higher level, techniques such as graph clustering and subgraph selection further enhance the reliability and consistency of the results.

\subsection{Physical Procedure Modeling}
The evolution of Earth-related events adheres to specific physical constraints. Typical DL models function as ``black boxes'', which lack interpretability and physical feasibility. By contrast, graph-based structures can be naturally linked to the physical processes, making them more resilient to over-fitting compared to simple statistical methods. Usually, integrating physical knowledge into GNNs can be achieved through the following approaches: 1) node attribute design~\cite{xiao2022dual, ge2021multi}, 2) graph structure design~\cite{teng202372, moshe2020hydronets}, 3) graph regularization~\cite{cachay2021world}, and 4) message passing procedure~\cite{seo2019physics}. 
\par
One straightforward approach is to attach potential influencing variables as additional node features, a skill prevalent in tasks involving complex interdependencies among variables~\cite{lam2022graphcast, li2022field}. Despite the added value auxiliary data can provide, it increases the computational burden. Thereafter, a remedy for this is to ground the physical model within the graph topology. Li \textit{et al.}~\cite{li2020hybrid} employ a graph search algorithm that rapidly traverses the sewer network, considering both the topology and physical processes governing the sewer system. Liu \textit{et al.}~\cite{liu2023physics} use the fast marching method given a geological model to estimate connectivity between injectors and producers for production forecast. Wang \textit{et al.}~\cite{wang2020pm2} take an advection coefficient as the activated projection of the wind vector on node pairs, together with node attributes covering physical, chemical, and biological reactions, to enhance the domain knowledge in graph learning. Seo \textit{et al.}~\cite{seo2019differentiable} summarize graph updating functions and corresponding physical equations, as shown in Table \ref{tab:differentiable}. Furthermore, the physical knowledge can guide the training procedure~\cite{mcbrearty2022earthquake}. Utilizing and acquiring invariances and equivariances within GNNs contributes to maintaining physics consistency and symmetry. This implies that a model trained on a specific graph for modeling physical procedures can effortlessly perform zero-shot transfers to any new, unseen graph characterized by entirely distinct relations~\cite{keriven2019universal, lino2022multi}.

\begin{table*}[!t]
    \caption{Differentiable functions can be used to update the node and edge features~\cite{seo2019differentiable}. $\phi^e$, $\phi^v$, $\phi^u$ define edge, node, and global update functions, respectively. $\nabla$, $\Delta$, $\text{div}$, $\text{curl}$ are the gradient, Laplacian, divergence, and curl on graphs, respectively.}
    \label{tab:differentiable}
    \centering
    \begin{tabular}{l l l}
    \hline
        Mapping & Updating function & Physics examples \\
        \hline
        Node $\rightarrow$ Edge & $e_{i,j}=\phi^{e}(v_i,v_j)=(\nabla v)_{ij}$ & $\nabla \phi = -E$ (Electric field) \\
        Edge $\rightarrow$ Node & $v_i = \phi ^v (e_{ij}) = (\text{div}~ e)_i$ & $\nabla \cdot E = \rho/\epsilon_0 $ (Maxwell's equ.) \\
        Node $\rightarrow$ Node & $v_i=\phi^v(v_i,\{ v_{j:(i,j)\in \mathcal{E}}\}) = (\Delta v)_i$ & $\Delta \phi = 0$ (Laplace's equ.) \\
        Edge $\rightarrow$ 3-clique & $c_{ijk} = \phi^c (e_{ij},e_{jk},e_{ki}) = (\text{curl}~e)_{ijk}$ & $\nabla \times E = 0$ (Irrotational E-field)\\
        \hline\hline
        First order & $v'_i = v_i + \alpha\phi^v(v_i, \{ v_{j:(i,j)\in \mathcal{E}}\}) = v_i + \alpha(\Delta v)_i$ & $\dot u = \alpha \nabla^2 u$ (Diffusion eqn.)\\
        Second order & $v_i^{''} = 2v'_i - v_i +c^2\phi^v(v'_i,\{ v'_{j:(i,j)\in\mathcal{E}}\}) = 2v'_i -v_i + c^2(\Delta v')_i$ & $\ddot u = c^2\Delta^2u$ (Wave eqn.)\\
        \hline
    \end{tabular}
\end{table*}

\subsection{Label Scarcity}
Annotating EO datasets is an exceptionally labor-intensive, time-consuming, and costly procedure. EO imagery often lacks distinct visual features, therefore it requires huge labeling efforts from experts. Furthermore, the volume of data collected by sensors is rapidly increasing. Adding to the challenge, labels can quickly become outdated due to the pronounced changes in targets and sensing conditions. For example, there is often a dearth of timely, relevant training data in post-disaster areas. In such scenarios, supervised learning models, which typically excel in settings where they are trained on comprehensive, well-labeled datasets, struggle to generalize to new, unseen scenarios. 
\par
Semi-supervised or unsupervised learning alleviates the dependency on label information in the EO domain~\cite{ren2021semi, wang2022self, liu2020semisupervised, bi2018graph, li2021manifold}. Label propagation by GNNs is a special case of transductive learning. The underlying principle of label propagation is based on the assumption that when nodes are connected by an edge, they are likely to share similar labels. Graph-based transductive learning or semi-supervised learning can achieve adaptive classification performance across various sensors, seasonal changes, and geographical locations~\cite{saha2022multitarget, zhao2022graph}. Such approaches are also beneficial in scenarios with missing values~\cite{guan2022recognizing} or when certain measurements are unavailable. Guan \textit{et al.}~\cite{guan2022recognizing} use unsupervised graph learning to deal with partially missing geochemical samples. Its improved adaptability paves the way for more robust industrial deployment, avoiding the whole system collapsing when a single station corrupts~\cite{kim2021graph}. In the evaluation of landslide susceptibility, Zeng \textit{et al.}~\cite{zeng2022graph} notice that the inherent heterogeneity of complex geoenvironmental settings leads to diverse landslide characteristics. Therefore, the GraphSAGE~\cite{hamilton2017inductive} algorithm is adopted to generalize the sampling approach to unseen areas through inductive learning.

%% file: sources/4_applications.tex
\section{Bring GNNs to EO applications}\label{sec:applications}
\begin{figure}[!t]
    \centering
    \includegraphics[width=1\columnwidth,trim={0cm 0.5cm 10.9cm 0cm},clip]{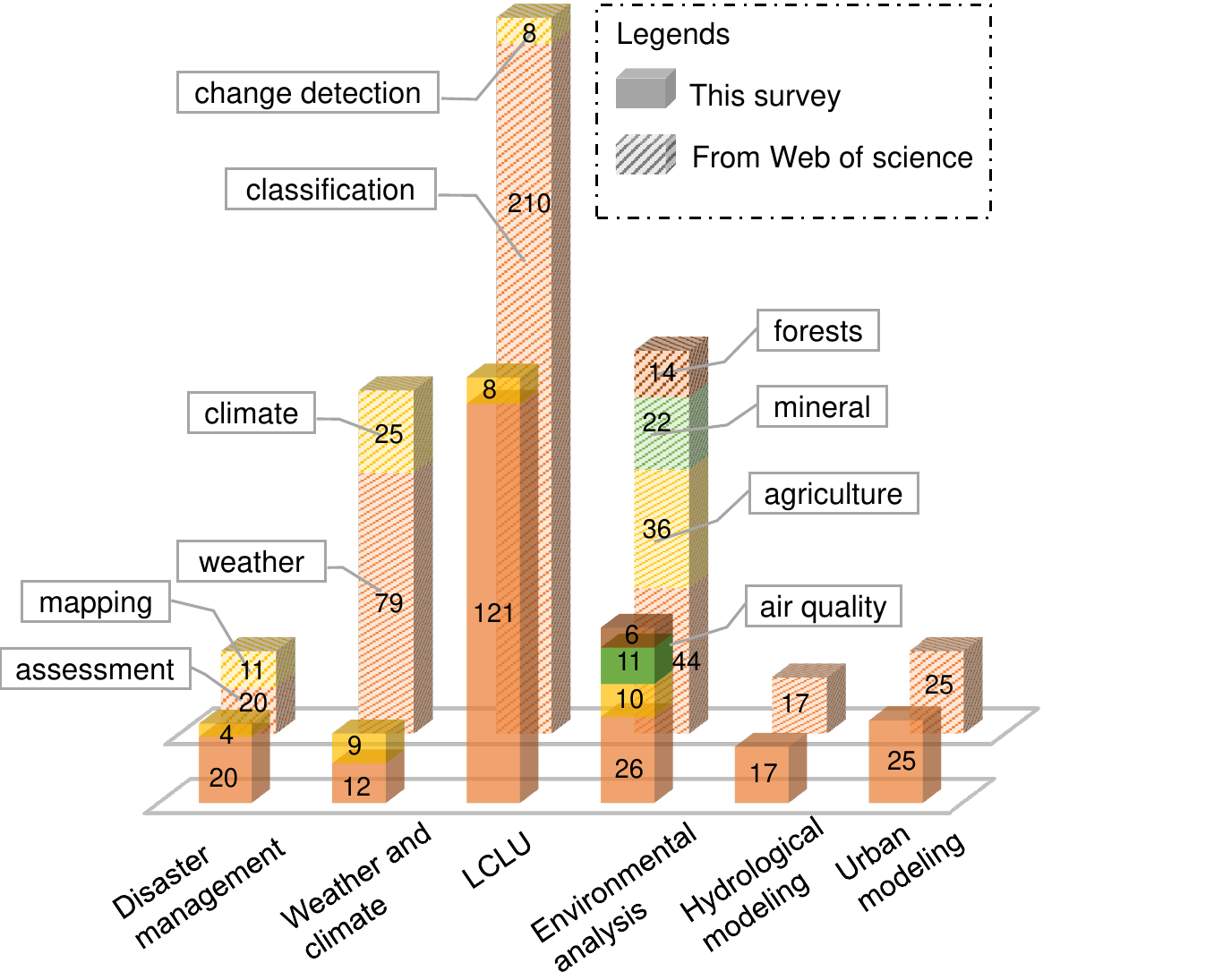}
    \caption{GNNs are widely adopted in various geoscientific research areas, though their applications are unevenly distributed. This survey selects key publications from the available resources, maintaining a balance similar to the original distribution.}
    \label{fig:number_papers}
\end{figure}

The advanced GNNs are important in a wide array of tasks such as predicting climate dynamics and meteorological patterns, offering insights for environmental analysis, managing and mitigating the impacts of natural disasters, accurate classification of land cover, and developing smart cities, etc. Figure~\ref{fig:number_papers} presents a relative estimate of the number of publications about GNNs within the EO domain. The data includes the number of papers gathered from the Web of Science\footnotemark\footnotetext{\url{https://www.webofscience.com}} and the number of high-quality papers selected for this survey. From this figure, it is evident that GNNs are particularly popular in topics related to Land Cover Land Use (LCLU), which are mainly based on hyperspectral image analysis. They are also extensively applied in air quality monitoring and weather forecasting tasks. This trend suggests that the data type largely drives the motivation for adopting GNNs, as these unstructured data (e.g., stations) requires new techniques to handle their irregularity.
\par
\re{In the subsequent section, we will delve into the key methodological considerations for designing and implementing GNNs across different EO applications, as well as their real-world implications.}

\subsection{Weather and Climate Analysis}
\subsubsection{Weather forecast}
% significance of weather forecast, conventional methods and their limitations
Weather forecasting and meteorological variable analysis, in particular severe and extreme weather, have always been the most important subject in meteorology. To predict the atmospheric conditions in the future, Numerical Weather Prediction (NWP) approximates the governing equations of fluid mechanics through discretization~\cite{kimura2002numerical}; but there remain considerable uncertainties~\cite{moosavi2021machine} due to the finite grid resolution and errors in parameterizing unresolved processes. Recently, DL has led to significant advancements in medium-range weather forecasts. These breakthroughs include extended forecast horizons, improved accuracy, reduced uncertainty, \re{much faster inference time, }and handling a larger set of variables. Conventional DL-based weather forecasting relies on latitude and longitude projections, which introduce singularities at the North and South poles. This results in an imbalance in training models due to the larger area along the equator compared to higher latitude areas. 
\par
\begin{figure}[!t]
    \centering
    \includegraphics[width=1\columnwidth,trim={0cm 1cm 0.8cm 1cm},clip]{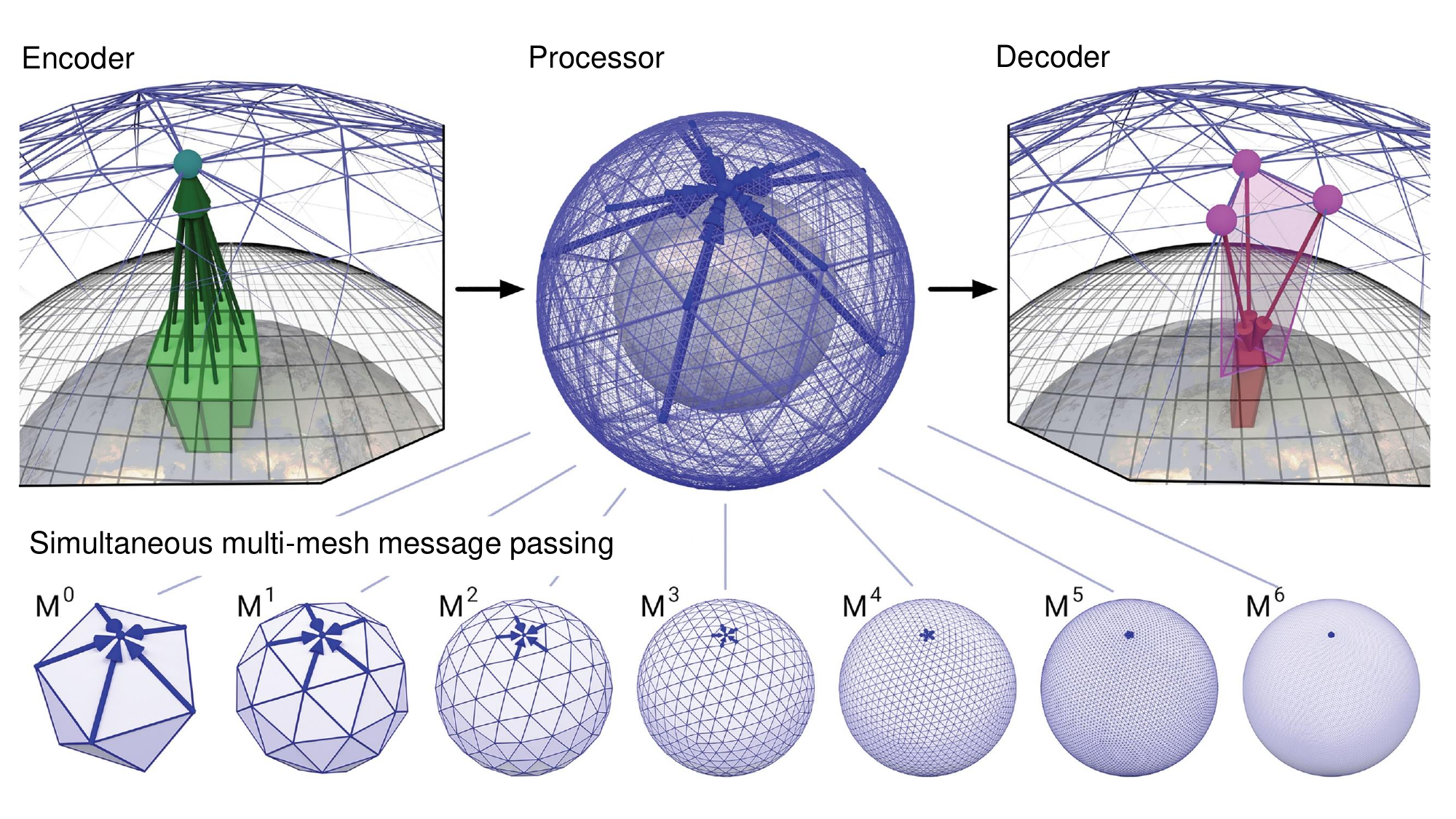}
    \caption{The encoder-processor-decoder architecture of GraphCast~\cite{lam2022graphcast} \textcopyright~\textit{2023, Lam \textit{et al.}, the American Associations for the Advancement of Science}. The encoder maps the latitude-longitude grided data to mesh representation. \re{The encoder contains a MLP that updates grid nodes without aggregation, and a Grid2Mesh GNN that updates mesh nodes by aggregating information from all ``grid to mesh edges" and updates ``grid to mesh edges" using information from adjacent mesh nodes and grid nodes; the processor (Multi-Mesh GNN) performs message-passing on the multimesh; and the decoder projects the mesh representation back to the original grided space. Specifically, the decoder is a Mesh2Grid GNN that updates grid nodes by aggregating information from all ``mesh-to-grid" edges and updates ``mesh-to-grid" edges using information from adjacent grid and mesh nodes.}}
    \label{fig:graphcast}
\end{figure}

% solution from gnns: 1) deal with spherical earth; 2) deal with scattered stations
To mitigate distortions arising from Earth's spherical geometry, Weyn \textit{et al.}~\cite{weyn2020improving} remap the Earth to an equiangular gnomonic cubed sphere and introduce convolutional and padding operations on it. While the Graph Network (GN) shows promising forecasting ability, it still failed to compete with NWP. Later, Keisler~\cite{keisler2022forecasting} use an encoder to map latitude-longitude grided data to an icosahedron. Their approach allows for aggregation on the sphere across physically uniform neighborhoods and enables adaptive mesh refinement. \re{It marked the first time that GNNs achieved Root Mean Squared Error (RMSE) values on Z500 (geopotential height at pressure level 500 meters) and T850 (temperature at pressure level 850 meters) comparable to those of operational, full-resolution physical models from Global Forecast System (GFS) and European Centre for Medium-Range Weather Forecasts (ECMWF). Building on this, GraphCast~\cite{lam2022graphcast}, whose graph organization is detailed in Table~\ref{tab:graphCast}}, extends the mesh representation to multiple scales by iteratively refining a regular icosahedron six times (\texttt{M\_0} to \texttt{M\_6}). \re{The 16 simultaneous message-passing layers transmit and aggregate spatial patterns over arbitrary ranges, and the nodes and edges are updated iteratively, }as depicted in Fig.\ref{fig:graphcast}. \re{Ultimately, GraphCast outperformed HRES (atmospheric model high resolution 10-day forecast) in skill scores across all lead times (1 day to 10 days) for Z500, with improvements around 7\% to 14\%. Its real-time forecasts are available on ECMWF’s OpenCharts\footnotemark \footnotetext{\url{https://charts.ecmwf.int/}}. }This GNN-based model overcomes the limitation of regularly strided ranges in CNNs and shows higher computational efficiency compared to the pairwise interactions computation in Transformers~\cite{vaswani2017attention}. Nevertheless, it faces some problems such as a lack of explicit incorporation of physical knowledge, deterministic outputs, and inherited bias from training data. To evaluate the added value of DL-based weather forecast for end users, more comprehensive protocols are necessary. Specifically, model performance should be validated by testing temporal consistency, the distribution of extreme events, and the capacity to preserve meteorological features, etc. 
\begin{table*}[htbp!]
    \centering
    \caption{\re{Components of GraphCast. The node features are composed of various data sources, such as forcing, static variables, and weather states. The edges connects neighboring nodes.}}\label{tab:graphCast}
    \begin{tabular}{p{1cm} p{2.2cm} p{4cm} p{1cm} p{7cm}}
    \hline
         \textbf{Unit} & \textbf{Type} & \textbf{Definition} &  \textbf{Numbers} & \textbf{Initial features} \\
        \hline
        \multirow{2}{*}{Nodes} & Grid nodes & Latitude-longitude point & 1,038,240 & \textbf{Weather states} $\times$ 2 time steps (6 atmospheric variables for 37 levels and 5 surface variables); \textbf{forcing} $\times$ 3 time steps (the total incident solar radiation at the top of the atmosphere, accumulated over 1 hour, the sine and cosine of the local time of day, and the sine and cosine of the of year progress); \textbf{static features} (the land-sea mask, the geopotential at the surface, the cosine of the latitude, and the sine and cosine of the longitude) \\ \cline{2-5}
        ~ & Mesh nodes & Uniformly placed points around a unit sphere (\texttt{M\_0} to \texttt{M\_6}) & 40,962 & Cosine of the \textbf{latitude}, and the sine and cosine of the \textbf{longitude}\\ \hline
        \multirow{3}{*}{Edges} & Mesh edges & Bidirectional connection between the mesh nodes for all scales & 327,660 & \textbf{Position} on the unit sphere of the mesh nodes (length of the edge, and the vector difference between the 3D positions of the sender node and the receiver node)
        \\\cline{2-5}
        ~ & Grid to mesh edges & Unidirectional edges from sender grid nodes to receiver mesh nodes within distance threshold & 1,618,746 & \textbf{Position} on the unit sphere of the mesh nodes \\  \cline{2-5}
       ~ & Mesh to grid edges & Unidirectional edges from sender mesh nodes to receiver grid nodes & 3,114,720 & \textbf{Position} on the unit sphere of the mesh nodes \\
        \hline
        \end{tabular}

    \end{table*}

\par
GNNs also demonstrate proficiency in handling meteorological observations from dispersed monitoring stations. For weather observations from weather stations, it is customary to conceive each station as a node, upon which meteorological measurements are affixed. The spectral convolution GNN~\cite{khodayar2018spatio, geng2021graph} utilizes temporal features from wind farms as nodes to forecast short-term wind speed. Given the important role that geography plays in shaping precipitation patterns, Huang \textit{et al.}~\cite{huang2022multigraph} organize clusters of ground radar stations as graphs based on their locations. Consequently, distinctive attributes characterizing diverse precipitation types are projected onto separate graphs. Similarly, Ma \textit{et al.} organize hierarchical graphs to model correlations between meteorological variables across weather stations (regions)~\cite{ma2023histgnn}.

\subsubsection{Climate analysis}
% climate analysis
In addition to atmospheric variables, Earth involves a myriad of other variables that engage in complex interactions. GNN-based learned simulators demonstrate promising performance in studying the complex physical dynamics of fluids and materials~\cite{sanchez2020learning, pfaff2020learning}, thus they have found applications in oceanic systems that show similar motion properties. 
\par
% gnn: 1) capture complex interaction, 2) improve process understanding
GNNs demonstrate remarkable predictability in complex systems dynamics, such as the prediction of sea surface temperature (SST)~\cite{wang2022time, ahmadi2020memory}, modeling of large ocean-circulation dynamics~\cite{cachay2021world}, and reconstruction of ocean deoxygenation~\cite{bin20244Dgraph}. To model the SST dynamics, Wang \textit{et al.}~\cite{wang2022time} use an adaptive graph trained from the dot product of node embeddings, indicating the pair-wise interactions in a high-dimensional feature space. Gao \textit{et al.}~\cite{gao2023global} superimpose a static $\mathcal{A}$ derived from pair-wise node similarity and a dynamic $\mathcal{A}$ composed of attention coefficients for SST prediction. This approach allows for the capture of adaptive node importance, as SST changes may be driven by signals originating from diverse regions of the ocean. Seo \textit{et al.}~\cite{seo2019physics} integrate transportation functions, primarily advection and diffusion, into GNNs for SST prediction. A multi-layer GN iteratively updates nodes and edges with global aggregation, during which the hidden state of the GN is constrained by diffusion equations, and the number of hidden layers corresponds to the order of spatial/temporal derivatives. 
\par
More importantly, GNN-based approaches offer opportunities to explore the information flow of underlying dynamic systems, potentially leading to scientific discoveries and improved process understanding. Cachay \textit{et al.}~\cite{cachay2021world} use GNNs to model large-scale spatial dependencies such as El Ni$\Tilde{\text{n}}$o events. This framework organizes SST and heat content anomalies for each longitude and latitude grid as nodes within a learnable graph structure. The learnable graph structure provides insights into the explicit information flow through edges across different time horizons. Despite such insights, they still confront challenges in terms of interpretability. This opacity can hinder our ability to decipher why and how certain predictions are made, and whether the information flow aligns with the physics.
\begin{figure*}[!t]
    \centering
\includegraphics[width=1.8\columnwidth,trim={0cm 0cm 4.6cm 0.6cm},clip]{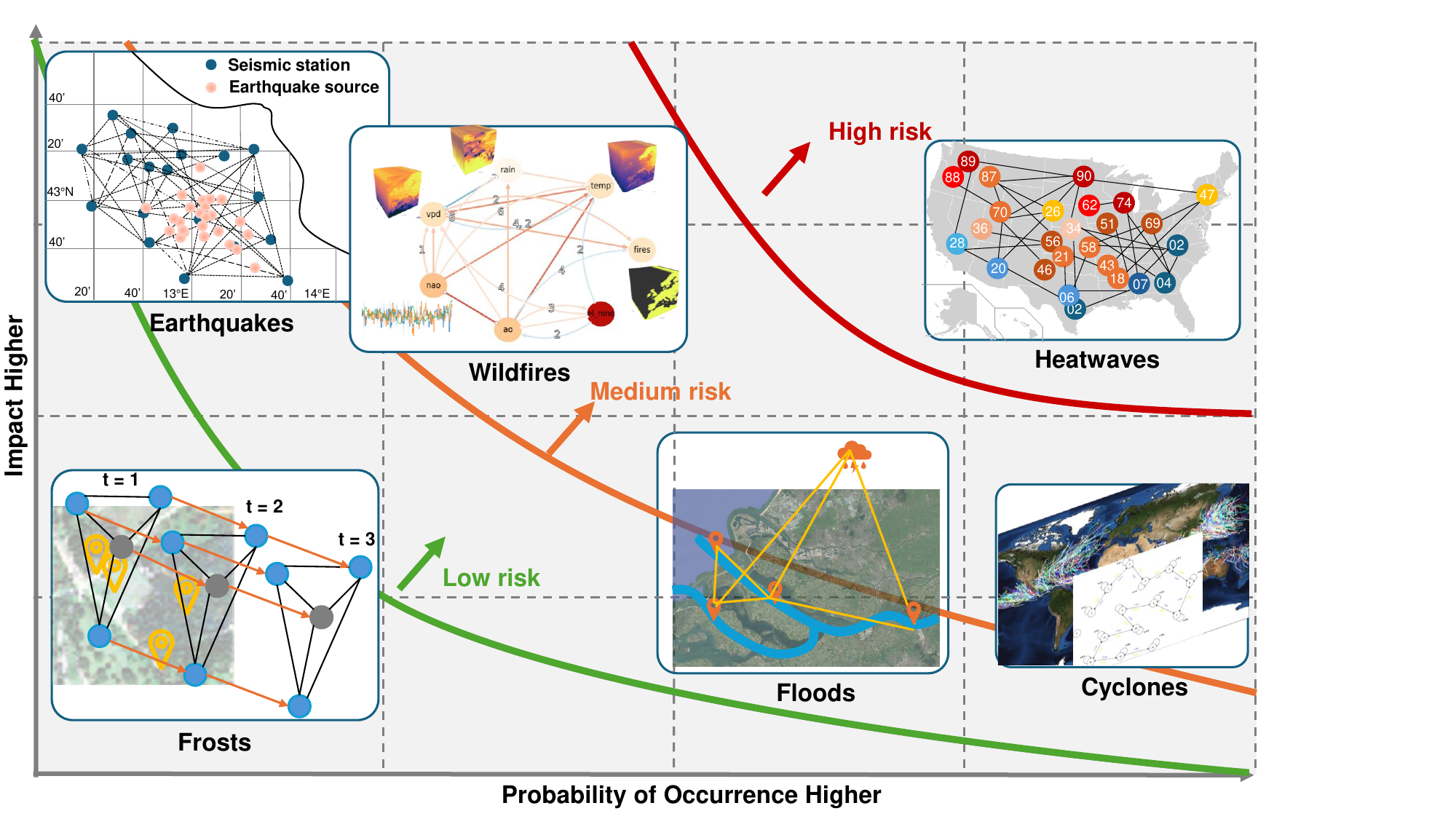}
    \caption{GNNs play roles in predicting, mapping, and assessing natural hazards, such as frosts~\cite{lira2021frost}, earthquakes~\cite{bloemheuvel2201multivariate}, floods~\cite{kazadi2022flood}, wildfires~\cite{zhao2024causal}, tropical cyclones~\cite{xu2023tfg}, and heatwaves~\cite{li2023regional}. The risk of the disaster is jointly determined by its probability of occurrence and impact. The accurate, reliable, and robust model prediction is crucial in managing high-risk disasters. The figure illustrates just one sample risk classification; the risk level for each disaster may vary depending on the specific area.}
    \label{fig:disasters}
\end{figure*}
\subsection{Disaster Management}
% impact of disaster, role of gnn in diaster management
Disasters and their cascading effects incur huge social and economic losses. Effective disaster management requires timely and precise disaster detection, including its characterization and temporal derivation. In this part, through examining a vast number of natural hazards such as landslides~\cite{li2023dci}, tropical cyclones~\cite{xu2023tfg}, floods~\cite{kazadi2022flood}, earthquakes~\cite{yano2021graph}, frost~\cite{lira2021frost}, and heatwaves~\cite{li2023regional}, we find GNNs are broadly adopted in all phases of disaster management, including pre-disaster prediction, mid-disaster mapping, and post-disaster reconstruction stage. Figure~\ref{fig:disasters} provides an overview of various natural hazards for which GNNs have been adopted for detection. The following capacities of GNNs make them particularly popular: 1) handling the irregular spatial arrangement of disaster observations~\cite{van2020automated}, 2) integrating a variety of factors as augmented node features~\cite{zeng2022graph}, 3) adjusting node density~\cite{jiang2022modeling} for rapid response, 4) extracting fine-grained features for detailed damage mapping, and 5) managing complex non-uniform layouts in post-disaster environments. 

\subsubsection{Disaster mapping}
% data used for diaster mapping
Successful monitoring and detection of potential hazards entail the use of various sensors, such as seismometers for seismic activity monitoring, weather stations for tracking meteorological conditions, and satellite imagery for remote sensing and observation. \re{GNNs effectively leverage these observations to enhance hazard detection, tracking, and monitoring. Their ability to map fine-grained features alleviates the scarcity of semantic information, enabling scene representations that better align with the cognitive needs of the public~\cite{li2022investigations}.}
\par
\noindent\textbf{Seismic activity:}
% Earthquake, conventional methods and their limitations
A sudden release of energy in the Earth's lithosphere creates seismic waves, which bring huge effect to human life and the natural environment. Seismic stations collect measurements for seismic event prediction, detection~\cite{yano2021graph}, classification~\cite{kim2021graph}, and source characterization~\cite{van2020automated, mcbrearty2022earthquake, bloemheuvel2201multivariate, zhang2022spatiotemporal}. Traditional methods treat this as an inverse problem, attempting to ascertain the location and magnitude of seismic events from arrival times or waveforms at a single station. However, these methods are computationally expensive and susceptible to signal noise. Hence, joint analysis of data from multiple seismic stations is widely adopted to reduce false detections~\cite{kim2021graph, yano2021graph} and identify even faint seismic signals.
\par
\begin{figure}[!t]
    \centering
    \includegraphics[width=1\columnwidth,trim={0cm 1.5cm 1.8cm 0cm},clip]{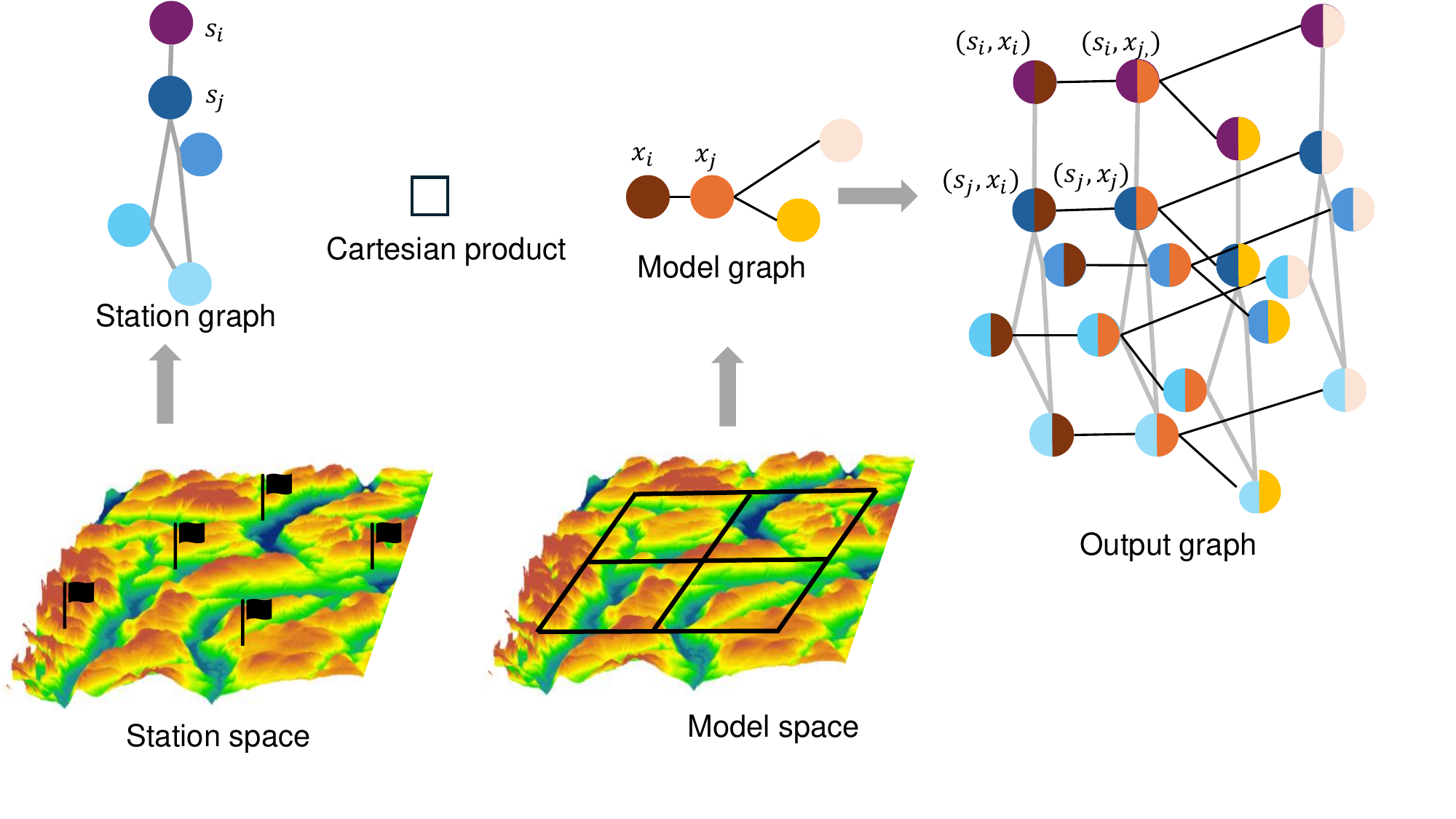}
    \caption{Cartesian graph of station graph and model graph. The nodes in the station graph are seismic stations, and the nodes in the model graph are evenly partitioned grids. The resulting graph preserves typologies of both the station graph and the model graph, and the resulting node attributes are a combination of both station nodes and model nodes. }
    \label{fig:cartesian}
\end{figure}

% solution from gnns: 1) handle irregular scattered stations
To address the irregular geometry of station networks and account for correlated noise between stations, graph regression and classification methods are employed. In related works, each seismic station is organized as a node in the graph, with graph connectivity determined by geographical proximity (threshold by distance or neighbor quantity)~\cite{kim2021graph, yano2021graph, bloemheuvel2201multivariate}, feature similarity, or a hybrid use of both~\cite{zhang2022spatiotemporal}. Nodes are attached with time-series waveforms processed by CNNs with varying kernel sizes or pooling techniques for temporal feature selection. Subsequently, GCNs or message-passing layers are used to encourage information exchange and aggregation across multiple sites~\cite{kim2021graph, bloemheuvel2201multivariate, mcbrearty2022earthquake}. In this context, the output is primarily graph-level attributes aggregated from node-specific features. Thereafter, concerns have arisen regarding potential feature loss incurred by such reduction operations~\cite{bloemheuvel2201multivariate}. To prevent information loss, Yano \textit{et al.}~\cite{yano2021graph} employ modularity maximization to cluster seismic station waveforms into groups. This clustering effectively accommodates the nonequispaced distribution of stations and enhances feature consistency within each group. The resulting subgraphs yield more representative statistics. McBrearty \textit{et al.}~\cite{mcbrearty2022earthquake} introduce a Cartesian product graph of station graph $\mathcal{S}$ and model graph $\mathcal{H}$, where $\mathcal{H}$ spans the entire searching space of the target predictands (magnitude and location). The resulting Cartesian product graph features two types of edges, as exemplified in Fig.\ref{fig:cartesian}. They provide complementary insights into local and global features between source/magnitude nodes and station nodes on the graph. The final prediction is obtained from the model space by stacking over stations processed after GCNs. Future research could be directed towards developing approaches to accommodate dynamic station numbers.

\par
\noindent\textbf{Meteorological and climatological disasters:}
% disaster prediction. solution from gnns: 1) handle multiple variables
The occurrence of natural hazards is seldom isolated, as they typically result from complex interactions and the collective influence of multiple contributing factors. Therefore, in disaster prediction, it is beneficial to incorporate a wide spectrum of signals to capture the potential of these events. Such signals commonly serve as augmented node features. For instance, FloodGNN~\cite{kazadi2022flood} utilizes scalar features such as ground elevation, Manning's friction coefficient, distance to the nearest river/stream, magnitude of velocities, as well as water depth to predict future water depths and velocities. Similar environments can give rise to comparable landslide conditions, even when they are geographically distant. Motivated by this observation, Zeng \textit{et al.}~\cite{zeng2022graph} compute the cosine similarity between environmental factors (e.g., drainage, faults, stratum, soil type) and unit characteristics (e.g., slope, aspect, normalized difference vegetation index, land use, and land cover) to quantify graph edge weights. For the prediction of regional heatwaves, Li \textit{et al.}~\cite{li2023regional} leverage a comprehensive set of node features, including temperature statistics, dew point temperature, wind speed, atmospheric pressure, ONI, day of year, and positional information. These factors encompass both high-frequency and long-term climatological effects, whose efficacy is validated through feature sensitivity.

\subsubsection{Disaster response}
% solution from gnns: 1) adjust ndoe density: fast response; 2) robust response
EO data assists disaster response by furnishing real-time and comprehensive information about affected areas. Quick disaster response, including monitoring spatial and temporal dynamics and rapidly mapping damage, is essential for smart rescue operations. Besides, natural hazards, such as landslides, typically result in significant alterations to the Earth's surface. This poses challenges for optical images due to drastic variations in illumination and environmental factors. In such circumstances, robust and reliable predictions are particularly vital for disaster management practices.
\par
GNNs offer the flexibility to adjust node density, enhancing inference speed. In the task of modeling wildfire spreads, Jiang \textit{et al.}~\cite{jiang2022modeling} adjust graph node density according to landscape complexity. The finer scale features over more complex regions are captured by the set of denser nodes, meanwhile, the model efficiency is improved by placing sparser nodes over simpler regions.
\par
In critical decision-making processes, GNNs can improve the robustness and reliability of model output by systematically organizing and aggregating information from multiple neighbors. SAR images are valuable for post-earthquake change detection, but they suffer from speckle noise. Wang \textit{et al.}~\cite{wang2021dynamic} split SAR images into overlapping blocks, treating superpixels within each block as nodes, and iteratively update node features by combining neighbor embeddings within the block. Although this design has proven robust to speckle noise, it is computationally intensive and its featuring scope is limited to local areas. Moreover, GNNs can adapt to heterogeneous areas. Berton \textit{et al.}~\cite{berton2023graph} address wildfire dynamics on a global scale by using a graph-based study of satellite imagery. They establish graph edges through a sliding window approach, leveraging correlations between successive events. Graph attributes are then employed for semi-supervised classification, providing insights into the spatial-temporal behavior of wildfires across diverse forest types and geographical regions.

\begin{table*}[!t]
        \caption{Prior knowledge and its integration into GNNs for air pollutant concentration prediction.}\label{tab:air_quality}
    \centering
    \begin{tabular}{p{5.5cm} p{6.5cm} p{4cm}}
    \hline
        \textbf{Factor} & \textbf{Relevance to targets} & \textbf{Integration into GNNs}\\
        \hline
        PM$_{2.5}$ & Air pollutant concentration presents spatial-temporal correlations & Spatial-temporal module \\
        \hline
        Other pollutants (e.g., PM$_{10}$, SO$_{2}$, O$_{3}$, NO$_{2}$) & Act as precursors or share similar atmospheric physical or chemical processes to the target pollutants & Node attributes\\
        \hline
        Aerosol Optical Depth & Provides additional air pollutant patterns & Node attributes\\
        \hline
        Meteorological factors (humidity, solar radiation, precipitation, planetary boundary layer height, K-index, temperature...) & Impact the formation, transmission, and accumulation of particles; crucial to long-term prediction & Node attributes, $\mathcal{A}$ (node-node similarity)\\
        \hline
        Wind speed & Impacts the transportation speed of pollutants & Edge attributes\\
        \hline
        Wind direction & Impacts the transportation direction of pollutants & Direction of the graph edge, edge attributes, advection coefficient\\
        \hline
        Point of interest & The land use and functions of the point (e.g., industry) is relevant to pollutant concentration & Node attributes\\
        \hline
        Road topology & Source of traffic pollution & Edge attributes\\
        \hline
        Longitude, latitude, altitude & Closer sites are more likely to have similar air pollutant concentration (First Law of Geography); Mountains block the transmission of air pollution & $\mathcal{A}$ (site-site connectivity)\\
        \hline
        Time (day of year, time of day) & Air pollutant concentration presents temporal patterns & Temporal edge direction, causal constraints, node attributes\\
        \hline
        Disputation function & Relates to temporal derivations of particle concentration & Constraints to the training of NNs\\
        \hline
        \end{tabular}

    \end{table*}
\subsubsection{Disaster assessment}
% significance of disaster assessment
After a disaster happened, assessing the damaged area is the first step toward restoring it to its pre-disaster state or better. More importantly, assessing the extent of damage can prioritize recovery efforts. Especially in complex urban environments characterized by non-uniform layouts, GNNs excel at capturing and representing the intricacies of the surroundings.
\par
% gnn: capture non-uniform layouts
Given that the majority of destruction typically clusters around the disaster's epicenter, neighboring buildings are likely to exhibit similar damage patterns. Ismail and Awad~\cite{ismail2022towards, ismail2022bldnet} use a graph-based approach to model spatial spreading dynamics in urban areas. In this framework, individual buildings are represented as nodes, equipped with features derived from building images that encapsulate their visual representations and feature differences before and after the disaster. Graph edges are established via Delaunay triangulation based on building envelope centroid pixel coordinates, with edge weighting based on spatial and spectral similarity, as proposed by Saha \textit{et al.}~\cite{saha2020semisupervised}. \re{The RS data provides essential and diverse features for producing reliable building damage maps~\cite{khankeshizadeh2024novel}. To evaluate the extent of building damage after a disaster,} Wang \textit{et al.}~\cite{wang2023hierarchical} try to answer fine-grained inquiries from coarser-grained questions. GNNs are then used to model hierarchical correlations among regions with different damage levels.
\par
To conclude, these studies exemplify the growing trend of using GNNs to enhance predictive capabilities for a range of natural disasters and environmental phenomena. GNNs have emerged as a potent framework by enabling the integration of diverse data sources, capturing spatial and temporal dependencies of data, and improving the generalizability and flexibility of model performance.
\subsection{Environmental Analysis}
% significance of environmental analysis, traditional methods and limitations
Environmental monitoring plays an important role in safeguarding ecosystems and human health by tracking changes in air quality, water resources, biodiversity, and others. Traditional CNNs may struggle to effectively capture relationships between sensor readings. GNNs offer a suitable solution by inherently accommodating irregular data structures, enabling accurate modeling of spatial dependencies, and facilitating robust predictions. This section delves into the applications of GNNs in air quality monitoring, forestry-related tasks, precision agriculture, and natural resources exploitation.
\subsubsection{Air quality monitoring}
% significance of air quality monitoring, motivation of using gnns
Air quality monitoring is closely related to human health, local ecosystems, urban administration, and sustainability development. Air pollutant concentrations are typically obtained from monitoring stations whose distribution is irregular and sparse, making them suitable for graph-structured data analysis. 
\begin{figure*}[!t]
    \centering
\includegraphics[width=1.8\columnwidth,trim={0cm 0cm 0cm 0cm},clip]{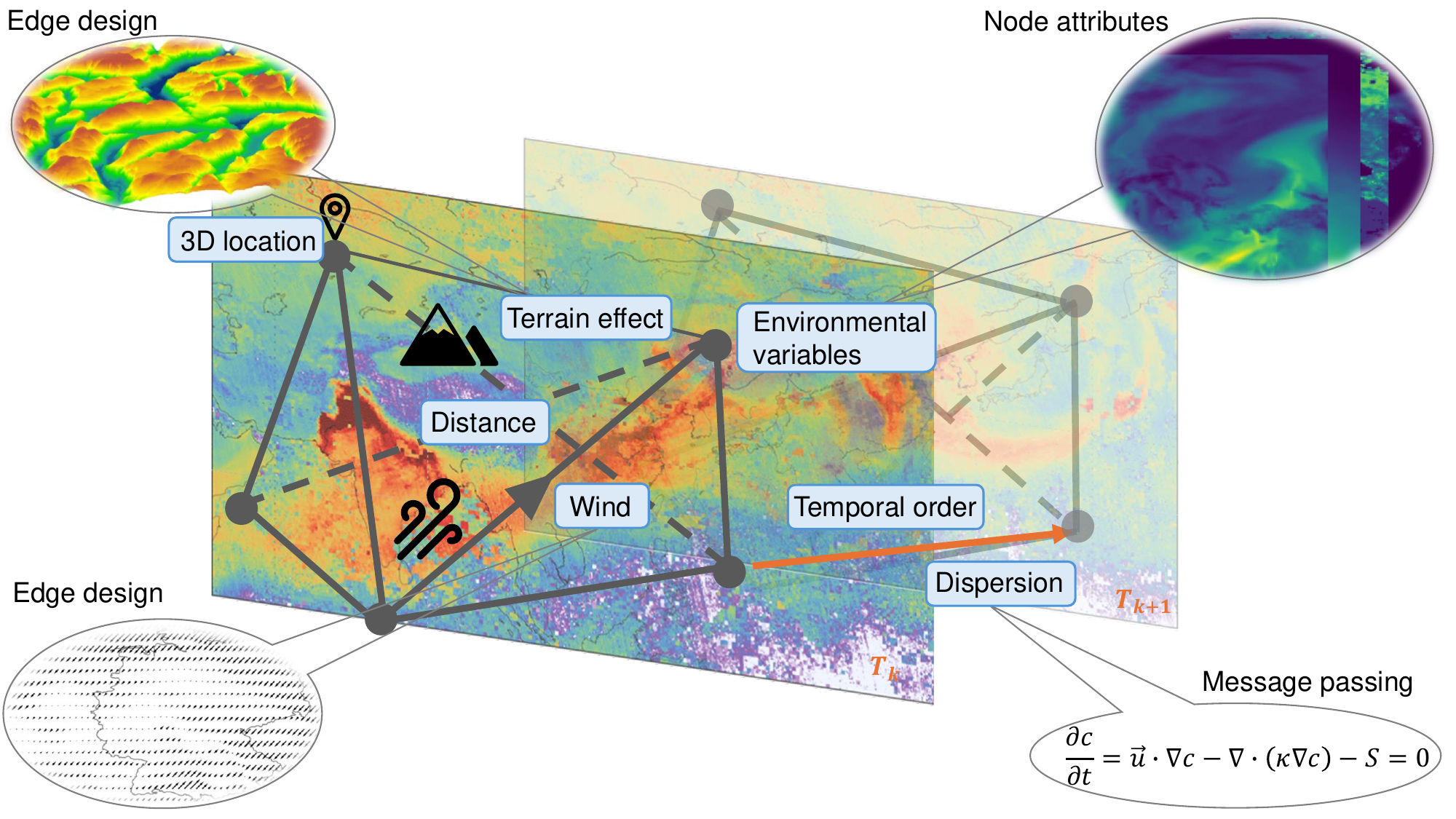}
    \caption{Abundant prior knowledge improves the GNN-based air quality prediction. In addition to the air pollutants measured at the stations, data from other sources such as DEM, reanalysis data, and RS imagery can be integrated into the graph design. Furthermore, the physical propagation functions of pollutants can inform and guide the graph learning process. Sample images are from Sentinel-5P, ECMWF Reanalysis v5 (ERA5), and Shuttle Radar Topography Mission (SRTM).}
    \label{fig:air_quality}
\end{figure*}
\par
% motivation 1) spatial-temporal dynamics: a.gnn+lstm; b.dynamic Adj; c.dynamic nodes
To capture the spatial-temporal dynamics of air pollutants, GNNs combined with temporal modules are widely adopted. There are various approaches to combine them. One simple yet effective solution is concatenating them in parallel. GC-LSTM~\cite{qi2019hybrid} takes the output of GNNs as input to Long Short-term Memory (LSTM) units, enabling the capture of both temporal and spatial correlations. However, its static $\mathcal{A}$ fails to encode the changing dynamics of these processes. Additionally, substantial distances between monitoring stations often limit the information that can get exchanged. An alternative approach involves employing a dynamic adjacency matrix. Xiao \textit{et al.}~\cite{xiao2022dual} additionally use wind speed and direction to organize a dynamic directed graph (DP-DDGCN). Then, a random walk Laplacian matrix is computed to simulate the diffusion process and capture complex contextual information. The adjacency matrix changes with current atmospheric conditions, and its temporal dynamics are captured by multiple gated recurrent units (GRUs). \re{The proposed DP-DDGCN model outperformed GC-LSTM across lead times up to 72 hours, with improvements in RMSE and mean absolute error (MAE) around 1\% to 4\%.} Teng \textit{et al.}~\cite{teng202372} simulate transportation paths in graph construction by considering geographical distance and elevation difference, with edge weights determined by wind information. \re{It not only achieved lower RMSE compared to GC-LSTM but also identified key source-receptor relationships.}
\par
% motivation 2) physcial processes
More importantly, the formation, transportation, and accumulation of air pollutants involve complex physical, chemical, and biological processes. The dynamic contextual information provided by these variables shows the effectiveness of GNNs in analyzing such complex procedures. For instance, considering the sensitivity of PM$_{2.5}$ prediction on domain knowledge, \re{PM{$_{2.5}$}-GNN~\cite{wang2020pm2} is designed with physical knowledge taken into account. It has been deployed online\footnotemark\footnotetext{\url{http://caiyunapp.com/map/}} to benefit the community. }Table~\ref{tab:air_quality} summarizes the variables that impact the concentration of PM$_{2.5}$ and the suggested approaches to integrate them into the model design. 
\par
% motivation 2) more reliable and robust prediction
Furthermore, the disruption or failure of monitoring stations and extending forecasting capability to new locations can be addressed by semi-supervised GNNs~\cite{tariq2023distance}. There are many efforts towards more stable prediction. For example, some researchers noticed the non-stationarity-related issues and the potential existence of pseudolinks. As GNNs are sensitive to noise and tend to aggregate improper global embeddings, improving the stability of graph embedding becomes a key concern. Mandal and Thakur~\cite{mandal2023city} propose separating the organized graph into several subgraphs based on the positional information and air pollutant concentration of nodes. This graph clustering method produces more robust forecast with fine spatial patterns.
\par
In summary, Fig.\ref{fig:air_quality} overviews various skills that can be integrated during graph learning. Despite the complex atmospheric processes considered, current research predominantly focuses on surface properties, often overlooking vertical interactions. Extending the analysis to 3D space could potentially further enhance predictive performance.

\subsubsection{Precision agriculture}
\begin{figure*}[!t]
    \centering
    \includegraphics[width=2 \columnwidth]{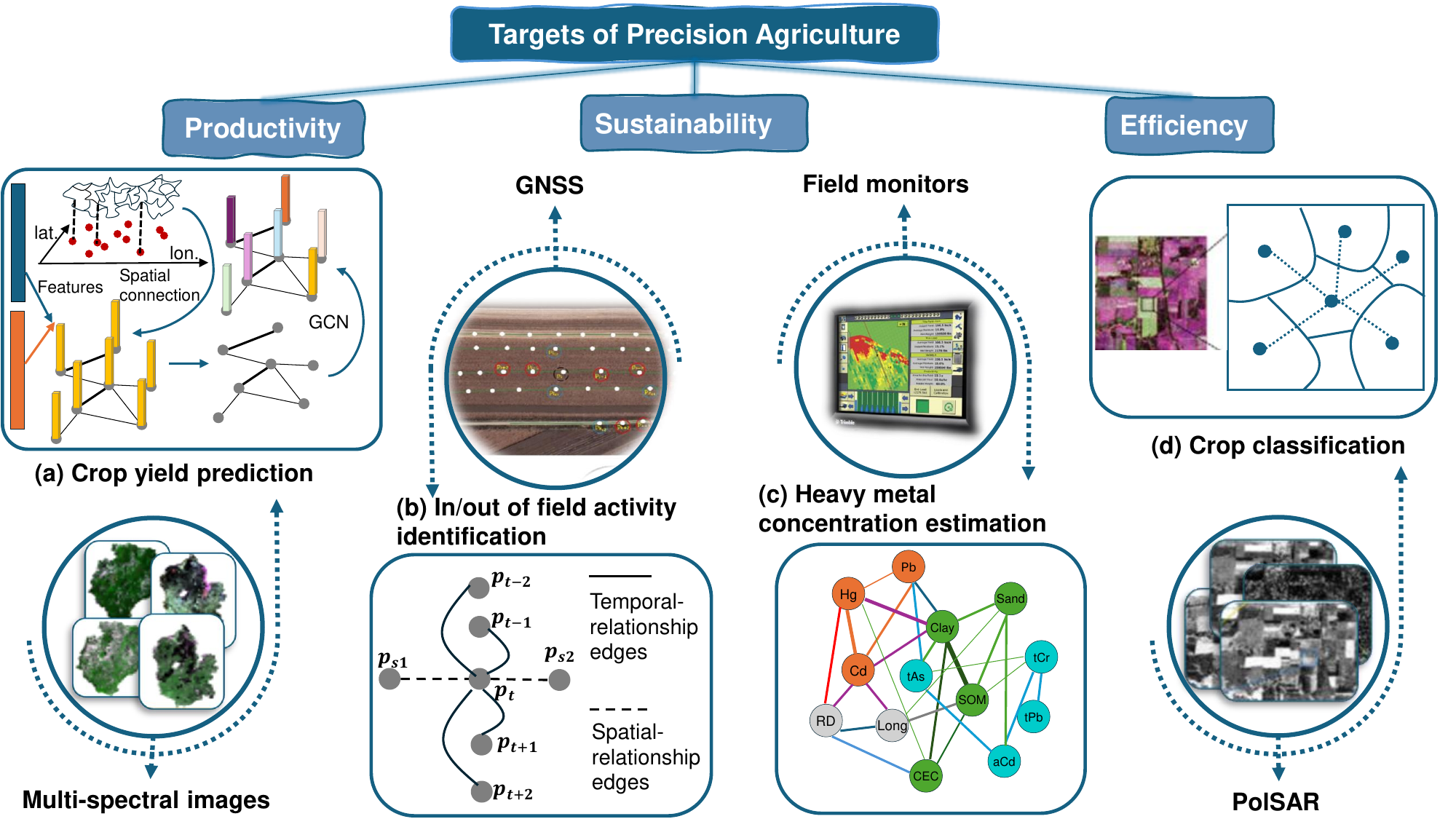}
    \caption{Applications of GNNs in precision agriculture: targets and tasks. (a) Time-series multi-spectral images are used to forecast crop yields~\cite{qiao2023kstage}. (b) Spatial and temporal neighboring GNSS points are connected to form a graph, enabling the identification of in-field and out-of-field activities~\cite{chen2022identifying}. (c) The chemical conversion processes of metal elements are modeled via a co-occurrence graph to estimate heavy metal concentrations in soils~\cite{li2022field}. (d) In PolSAR imagery, superpixels serve as nodes, while the feature similarity matrix acts as the adjacency matrix for crop classification~\cite{cheng2022novel}.}
    \label{fig:agri_applications}
\end{figure*}

% significance of precision agriculture
Precision agriculture (PA) involves collecting, modeling, and managing the temporal and spatial variability of crop fields to optimize agricultural efficiency, productivity, and sustainability~\cite{preciseAg}. EO data from RS (satellite, aerial, unmanned aerial vehicles (UVAs), etc), meteorological stations, GNSS, and sensory measurements provide abundant information to support the analysis, assessment~\cite{zhang2022suitability}, management, and economic planning of farming products, especially in the context of climate change. GNNs have been applied across various agricultural tasks, as illustrated in Fig.\ref{fig:agri_applications}.
\par
% Crop yield prediction. GNN: 1) empirical knowledge 2) spatial and temporal correlation
Crop growth is influenced by a multitude of complex environmental factors. \re{Therefore, in crop yield prediction, expert knowledge in agriculture, such as empirical insights into the importance of growth stages~\cite{qiao2023kstage}, the dispersion and diffusion processes of climate variables, and the geographical topology of stations, can be incorporated into the design and learning of graph representations}. This incorporation enhances the predictability and interpretability of the model. For instance, Qiao \textit{et al.}~\cite{qiao2023kstage} take a multi-head self-attention network to learn the temporal adjacency matrix. The temporal attention weights are artificially increased on heading and booting stages considering their greater influences on the final winter wheat yield. Besides, non-independent and identically distributed (i.i.d) crop data typically exhibit strong spatial and temporal correlations with high variability. \re{Experimental results demonstrate that the prior knowledge distillation into temporal networks improves the coefficient of determination ($R^2$) values for corn yield predictions by approximately 2.6\%. }To tackle the challenges posed by these complex and dynamic spatial-temporal correlations, Fan \textit{et al.}~\cite{fan2022gnn} utilize the output of GraphSAGE at each time step as the hidden state of RNNs. Graph connectivity is encoded using adjacent counties, and input features of GraphSAGE are derived from weather data, land surface characteristics, crop productivity indices, and soil quality, all of which impact the final yield. 
\par
% intelligent agriculture management. GNN: 1) handle high dimensional data
For intelligent agriculture management, activity analysis, crop type classification, and pest control are important topics to optimize operation plans and reduce crop costs. Irregular GNSS trajectory points and station measurements naturally lend themselves as suitable candidates for graph-based structures, as highlighted by Chen \textit{et al.}~\cite{chen2022identifying}. In~\cite{cheng2022novel}, time-series polarimetric SAR (PolSAR) data is used for crop classification. A tensor similarity value is devised to help find the global optimal neighborhoods within the entire SAR image. Consequently, the aggregated and transformed information of nodes by GCN provides deeper features from a global perspective, outperforming raw polarimetric features in terms of efficiency. Another example of a complex system is the soil-rice system~\cite{li2022field}, where the transmission of heavy metals involves processes that are not fully understood. GCN ignores the order of input factors, and the aggregation of neighborhood information simulates the influence diffusion among variables globally. Thus, GNNs have proven to be powerful tools for understanding high-dimensional and interconnected ecological data.

\subsubsection{Natural resource exploitation}
% significant of smart natural resources exploitation, challenge
Natural resources such as minerals and petroleum are vital to support industrial and information civilization. The collected EO data and its analysis can assist to allocate resources more effectively and predict the production. Smart resource exploitation not only mitigates economic and human labor wastage but also encourages environmentally conscious resource assessment and sustainable development. Nevertheless, the distribution of natural resources is often sparse and irregular, and their development is not only related to historical data at this position, but also to complex spatial dependencies among other sites and many environmental factors. 
\par
% gnn solution 1) complex variales
Therefore, GNNs have been applied to model these complex interdependencies among variables. In Table~\ref{tab:natural_resources}, we summarize the tasks and prior knowledge considered during graph construction. For instance, Zuo \textit{et al.}~\cite{zuo2023graph} delineate prospecting areas for minerals by considering features like northeast (NE-) and northwest (NW-) trending faults, Agno Batholithic pluton margins, and porphyry intrusive contacts, all of which exhibit strong spatial associations with gold mineralization. Quan \textit{et al.}~\cite{guan2022recognizing} observe that mineral deposits are often found in locations inconsistent with overall patterns. Inspired by this, they approach this task as an anomaly identification problem within land cover data. The distance-based adjacency matrix is thresholded at Moran's I tipping point to balance spatial structure information and spatial heterogeneity. The GAT is used to quantify the importance of faults and effectively aggregate information from all subspaces. To assess land degradation after mining, Zhou \textit{et al.}~\cite{zhou2023deep} embed the GCN with a CNN extracted adjacency matrix into the bottleneck of an auto-encoder. As for production prediction, Gao \textit{et al.}~\cite{gao2022production} consider factors such as the spatial distribution of well networks, gas injection data, and historical production records to predict multi-well production levels.
\begin{table}[!t]
        \caption{Prior knowledge (factor) and its integration into GNNs for natural resources exploitation.}\label{tab:natural_resources}
    \centering
    \begin{tabular}{p{3cm} p{2.5cm} p{1.5cm}}
    \hline
        \textbf{Factor} & \textbf{Application} & \textbf{Integration to GNNs}\\
        \hline
        Faults (NE, NW), Agno Batholithic pluton margins, porphyry intrusive contacts & Gold mineralization exploitation & Node attributes \\
        \hline
        Faults anomalies & Mineral exploitation & Edge weights \\
        \hline
        Wells location & Production forecasting &  $\mathcal{A}$ \\
        \hline
        Travel time between injectors and producers  & Hydrocarbon production forecasting &  $\mathcal{A}$ \\
        \hline
        Gas injection rate & Petroleum production prediction & Dynamic node attributes\\
        \hline
        \end{tabular}

    \end{table}

\subsubsection{Forest monitoring}
% Forest monitoring. gnn solution: 1) handle point clouds
Forest monitoring involves a broad of tasks aimed at assessing, managing, and preserving forest ecosystems. LiDAR point cloud data produces an accurate 3D representation of objects, and it can penetrate vegetation and foliage, allowing to capture ground-level data~\cite{li2021high} even in dense forests. Consequently, it serves as an important data source for forest type classification~\cite{guan2015deep}, Digital Elevation Models (DEM) extraction~\cite{zhang2020extraction} over forest area, and forest strata mapping~\cite{arvidsson2021predicting}, etc. GNNs are particularly well-suited for handling point cloud data, owing to their innate capacity to capture relationships among individual data points within unstructured data formats. In forested regions, collecting ground points via airborne laser scanning (ALS) can yield an extremely sparse dataset. To extract the DEM over the forest, Zhang \textit{et al.}~\cite{zhang2020extraction} propose Tin-EdgeConv, which considers the spatial relationship between ground and non-ground points. The method utilizes k-nearest neighbors (KNN) solely in vertical and horizontal directions to construct a directed graph. Other types of data, such as aerial images, also find applications in this field. Pei \textit{et al.}~\cite{pei2023application} embed the GCN between an encoder and a decoder. While the encoder uses CNNs for local feature capturing, the GCN learns the long-range information. \re{The proposed model achieved a kappa ($\kappa$) of 0.78 and an overall accuracy (OA) of 0.85 in forest type classification, outperforming the GCN’s $\kappa$ of 0.75 and OA of 0.82.}

\subsection{Land Cover Land Use}
% significance of LULC; limitation of convential method, emerging solution from gnns
Land Cover Land Use (LCLU) related tasks hold significant importance in the RS community because they provide fundamental input for interpreting RS imagery. CNNs often struggle to generate highly detailed features because of their inherent downsampling process. To deal with this limitation, GNN-based methods have been applied in several LULC-related tasks, including semantic (pixel-wise classification) and change detection. Probabilistic graph models, such as conditional random field~\cite{liu2015semantic}, have been used in this context. However, it hasn’t sufficiently extracted the features from the images and the propagation of information by using conventional graph models is not adequate. Hence, GNN-based methods, such as gated GCN~\cite{shi2020building}, which integrates the deep structured feature representation and the GCN, have been proposed. To extract  representative features with multi-level characteristics, the deep structured feature embedding technique is utilized for enhanced feature fusion. GCN with GRUs takes into account both local and global contextual dependencies for information propagation. More recently, GNN-based methods have focused on further architecture modifications, e.g., graph-in-graph network~\cite{jia2022graph}, self-constructing GNN~\cite{liu2021self}, and modifications in graph pooling~\cite{sarpong2023hyperspectral}.
\par
%  pixel-wise hyperspectral image classification
Hyperspectral images, known for their fine spectral resolution at the cost of spatial distinguishability, are notoriously hard to analyze due to their high dimensionality. GNNs aid in pixel-wise classification by capturing long-range dependencies and contextual information~\cite{mou2020nonlocal}. They also offer flexibility in integration with other modules such as CNNs and attention mechanisms~\cite{dong2022weighted}. Wan \textit{et al.}~\cite{wan2019multiscale} introduce a multi-scale dynamic GCN that updates the graph during training at various scales. Mou \textit{et al.}~\cite{mou2020nonlocal} propose a semi-supervised non-local GCN to facilitate global context modeling. Building on the advancements in attention mechanisms,~\cite{tong2020few, sha2020semisupervised} investigate graph networks with attention mechanisms for tasks related to hyperspectral image classification. Furthermore, hyperspectral image classification~\cite{ding2022multi} is explored in~\cite{he2021dual} through a combination of point graphs and distribution graphs. Although uncertainty quantification in hyperspectral image classification is less studied, some efforts have been made to address this gap. Yu \textit{et al.} develop evidential GCNs and graph posterior networks to estimate uncertainty in node classification~\cite{yu2023uncertainty}.
\par
% change detection
In addition to pixel-wise classification, pixel-wise change detection methods have also incorporated GNN-based architectures. In~\cite{saha2020semisupervised}, a semi-supervised approach is proposed relying on the presence of a limited quantity of multi-temporal labels denoting change and non-change. They introduce an innovative graph construction method designed to efficiently encapsulate spatial and spectral information across multiple scales. Similarly, Tang \textit{et al.}~\cite{tang2021unsupervised} combine GCN with metric learning for unsupervised change detection.  
\subsection{Hydrological Modeling}
% hydrology
Hydrological modeling, which includes rainfall-runoff hydrological process understanding and hydrodynamic process understanding, presents challenges due to the non-linearity of these processes. GNNs have demonstrated efficiency in capturing the spatio-temporal variability within such complex hydrological processes.
\par
% hydrology modeling. gnn 1) river topology; 2) physical equations
For hydrological modeling, the spatial variability of rivers often serves as a focal point in GNN design. Sit \textit{et al.}~\cite{sit2021short} develop a model based on graph convolutional GRUs to account for the spatial diversity of rivers. Moshe \textit{et al.} build HydroNets~\cite{moshe2020hydronets} upon river network structure. In this directed graph, each node signifies a basin, and the direction of an edge follows the water flow from a sub-basin to its containing basin. Jia \textit{et al.} propose a recurrent GNN~\cite{jia2021physics} to extract interdependencies among segments within river networks. The physics of streamflow is also integrated into the learning procedure. Moreover, hydrodynamic processes adhere to some fundamental laws of physics. This offers the chance to integrate these physical equations into GNN architecture design. Ma \textit{et al.} propose a hydrodynamic interaction GNN (HIGNN)~\cite{ma2022fast} to predict the particles’ dynamics in Stokes suspensions. While GNN-based models excel in capturing spatial-temporal dynamics, they also entail increased complexity. Future endeavors should prioritize enhancing the scalability and robustness of GNN-based approaches, particularly in modeling large watershed areas~\cite{xu2023physics}. 

\subsection{Urban Modeling}
% Urban characteristics
Urban scenes are composed of interdependent functional components, which necessities graph-structured representations. To develop digital twins of the built environment, these components need to be modeled as digital counterparts~\cite{borrmann2023artificial}. GNNs are particularly effective in urban modeling because of their ability to interpret and process complex, irregular urban structures.

% Urban point cloud understanding
\re{
Point clouds capture the built environment with a high level of detail, obtained from sources such as Photogrammetry, laser scanning, or TomoSAR. However, this high level of detail presents challenges in processing the data~\cite{xie2020linking}. A point cloud can naturally be conceptualized as a graph, where adjacency is determined by the spatial proximity of points through aggregation. Modern point cloud backbones can implicitly form graphs from these points~\cite{qi2017pointnet,qi2017pointnet++,wang2019dynamic}. Alternatively, the graph structure can be explicitly defined on a set of oversegmented points through geometric partitioning, commonly referred to as superpoints, analogous to superpixels in the image domain. This compact yet context-rich representation can then be leveraged by a GNN for semantic segmentation~\cite{landrieu2018large}. The superpoint representation has also been extended in subsequent works~\cite{robert2023efficient,robert2024scalable} for scalable semantic and panoptic segmentation of urban scenes.
}

% Layout modeling
Urban layout design involves creating complex city structures that balance aesthetic, functional, and spatial requirements. Xu \textit{et al.}~\cite{xu2021blockplanner} introduce a generative model for city blocks that utilizes a vectorized representation with a ring topology and a two-tier graph. This structure effectively captures both the global and local aspects of city blocks, facilitating the generation of diverse land lot configurations. It abstracts each land lot into a vector that includes 3D geometry and land use semantics, thereby aiding in urban planning. Similarly, He and Aliaga~\cite{he2023globalmapper} employ graph attention networks to generate realistic urban layouts given arbitrary road networks, offering versatility in representing different building shapes and handling complex layouts across cities. \re{They further introduce a graph-based autoencoder for improved realism and semantic consistency~\cite{he2024coho}.}

% Road modeling
For city road layout modeling, Chu \textit{et al.}~\cite{chu2019neural} introduce a generative model for spatial graphs, specifically focused on city road layout modeling. This model uses graph representation where nodes signify control points and edges represent road segments, iteratively generating new nodes and edges based on the current graph. He \textit{et al.}~\cite{he2020sat2graph} propose a new encoding scheme which allows for the representation of road graphs in tensor format. This approach simplifies the training of a non-recurrent, supervised model that can predict a rich set of features to capture the graph structure from a satellite image. Shit \textit{et al.}~\cite{shit2022relationformer} propose a one-stage transformer-based framework for joint prediction of objects and their relations in images. The architecture overcomes the limitations of traditional two-stage approaches by efficiently modeling object interactions and relationships to generate image-to-graph representations.

% 2D building (footprint) modeling
In the field of building footprint extraction, Shi \textit{et al.}~\cite{shi2019building} propose an end-to-end framework using a GCN for building footprint extraction. Zhang \textit{et al.}~\cite{zhang2020conv} design a message passing neural network to reconstruct an outdoor building as a planar graph from a single RGB image. This architecture is specifically tailored for scenarios where nodes of a graph have explicit spatial embedding, with nodes corresponding to building edges in an image. Zhao \textit{et al.}~\cite{zhao2022extracting} extract vectorized building rooflines and structures from high-resolution RS imagery, where building corners are represented as graph nodes. Their approach combines primitive detectors and GNN-based relationship inference. Likewise, Zorzi \textit{et al.}~\cite{zorzi2022polyworld} extract building vertices from images and accurately connect them to form precise building polygons. This method employs a GNN to predict the connection strength between vertex pairs and uses a differentiable optimal transport problem to estimate the assignments. Zorzi \textit{et al.}~\cite{zorzi2023re} further improve~\cite{zorzi2022polyworld} by featuring a redesigned architecture that leverages both vertex features and the visual appearance of edges. The edge-aware GNN efficiently predicts connections between vertex pairs, making it versatile for various applications like building extraction, floorplan reconstruction, and wireframe parsing. Wang \textit{et al.}\cite{wang2023regularized} employ a concept called primitive graph as a unified representation of vector maps. Their multi-stage learning process reconstructs primitive graphs, which enhances shape regularization and topology reconstruction capabilities.

% 3D building modeling
Going one dimension higher, 3D reconstruction poses even more challenges due to inherently error-prone topological constraints. A typical approach to urban modeling is the use of surface meshes~\cite{gao2021sum}. In boundary representation (B-rep), a surface mesh consists of facets, which are further composed of edges and vertices. \cite{gao2023pssnet} organizes the geometric and photometric features of the segments as graph nodes and the multi-level contextual features in graph edges, and performs mesh segmentation with a planarity-sensible GCN. To reconstruct compact building models, Chen \textit{et al.}~\cite{chen2022reconstructing} use an MRF to balance the fidelity and compactness of building models generated from photogrammetric point clouds, and solve it with graph cuts. They~\cite{chen2023polygnn} further utilize a polyhedron-based approach to assemble primitives obtained through polyhedral decomposition from airborne LiDAR. The reconstruction is then formulated as a graph node classification problem. Unlike mesh reconstruction, Jiang \textit{et al.}~\cite{jiang2023extracting} approach the wireframe of building models with a GNN embedded with corner information. Additionally, for instance-level building modeling, GNNs have found extensive applications in the construction domain as well~\cite{jia2023graph}.

\begin{figure*}[!t]
    \centering
    \includegraphics[width=1.1 \columnwidth]{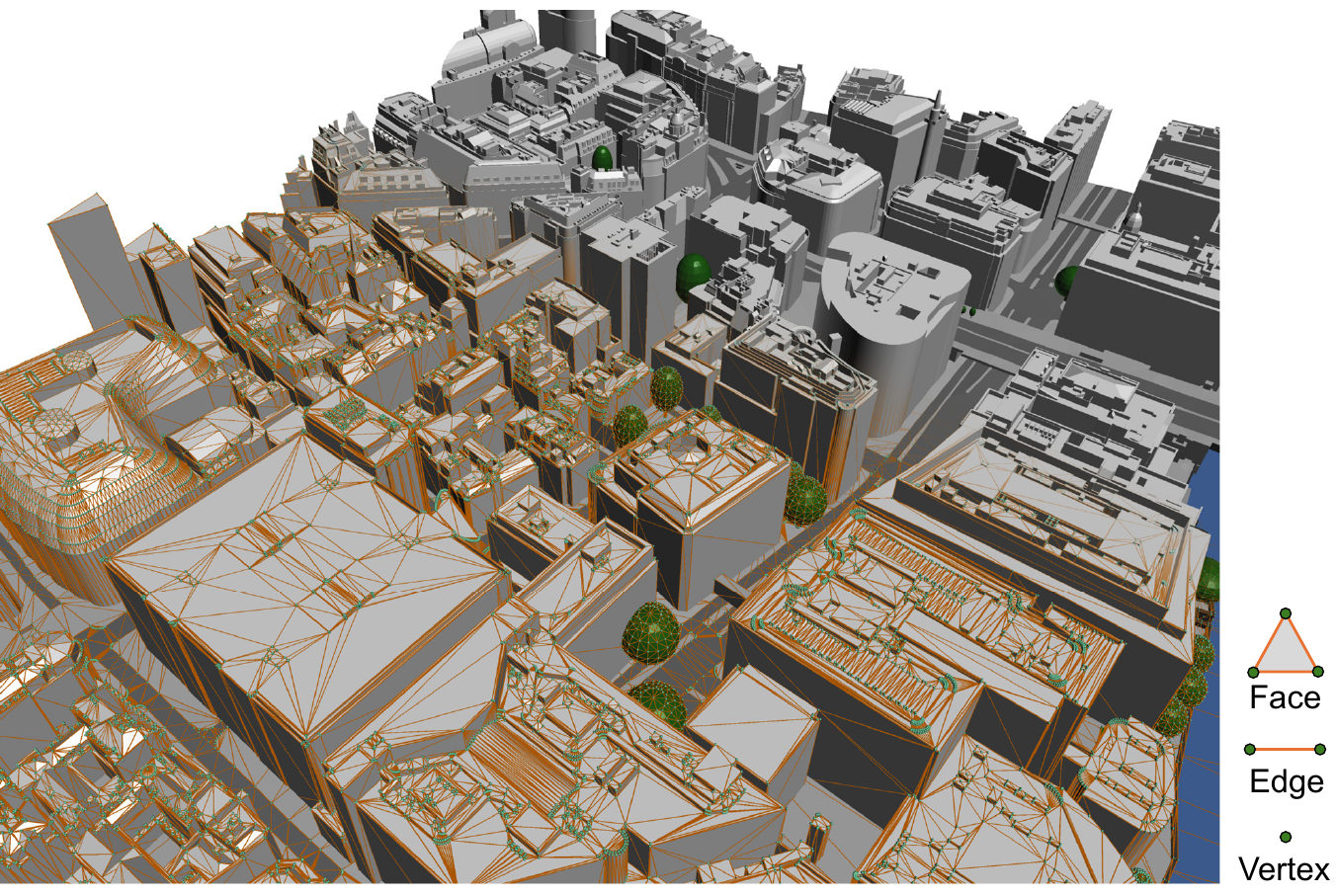}
    \caption{\re{Surface mesh of an area in London. In boundary representation, a surface mesh consists of faces, which are further composed of edges and vertices. GNNs can process this boundary representation data structure with graph representations.}}    \label{fig:urban_mesh}
\end{figure*}

%% file: sources/5_challenges.tex
\section{\re{Challenges of implementing GNNs}}
\label{sec:challenges}
Despite the merits of GNNs in addressing EO challenges, as discussed in the previous chapter, their adoption in the EO domain remains relatively underexplored. This discrepancy primarily arises at various stages, including data formatting, graph design, subsequent graph learning, and deployment to large-scale scenarios. Addressing these methodological challenges would pave the way for developing innovative approaches to analyze and interpret the Earth.

\subsection{Low-level Inputs}
Most GNNs are typically developed for graph-structured data. In other fields like chemistry, defining a graph is straightforward; for instance, atoms can serve as nodes, molecular bonds as edges, and molecules as graphs. This ease of definition has led to the widespread adoption of GNNs in biochemistry science, such as protein-to-protein prediction. However, many EO datasets lack inherent node definitions.
\par
The primary challenges revolve around extracting relevant variables from frequently noisy measurements or high-dimensional spatio-temporal datasets. To understand the scene depicted in an RS image, recognizing semantic relationships between pixels is important. This motivates the organization of pixels or superpixels as nodes. Another strategy entails mapping gridded measurements onto a spherical space, where the data distribution is more uniform and better aligned with Earth's physical attributes~\cite{lam2022graphcast}. Furthermore, dimensionality reduction can be used to extract meaningful nodes from these lower-level inputs. To detect tropical cyclone intensity from satellite images, Xu \textit{et al.}~\cite{xu2023tfg} conceive a specific wind scale as a node, with its vector attribute computed from the mean wind speed associated with that scale from all training data.
\par
Beyond the data domain, node definition can extend to other spaces. At the task level, Wang \textit{et al.}~\cite{wang2023hierarchical} exemplify this by organizing the three-level subtasks 
$-$ 1) whether it is a building; 2) whether the building is damaged; 3) to what extent the building is damaged (minor, major, destroyed) $-$ as three types of nodes in a tree-structured graph. This hierarchical modeling approach effectively breaks down the intricacies of the overall task.

\subsection{Implicit Graph Structure}
While most EO data is organized in a regular format, implementing GNNs involves crafting graph structures from data lacking inherent graph topology. 
\par
When designing the graph topology, the choice hinges on the specific scientific research question at hand. Geographical, environmental, and physical processes can be incorporated into the graph topology design. The configuration of the graph presents itself through an array of possibilities: one may base it on geographical proximities~\cite{qi2019hybrid}, exploring the mutual information between node pairs~\cite{khodayar2018spatio}, feature similarities~\cite{li2022field}, or embrace a trainable paradigm~\cite{geng2021graph, wilson2018low, cachay2021world}. The direction of edges defines the desired information flow, and often under certain constrictions to align the graph with reality. For example, Qiao \textit{et al.}~\cite{qiao2023kstage} enforce temporal edges to direct toward the future. Yang \textit{et al.} use an attention based directed graph generator for automatic vector extraction from RS images~\cite{yang2023topdig}. Recently, there has been growing interest in eliminating superficial edges by studying the underlying data causality for trustworthy GNNs~\cite{jiang2023survey}. In line with this, Zhao \textit{et al.} \cite{zhao2024causal} explore a causality-based adjacency matrix to capture the synergistic effect of variables driving wildfires.
\subsection{Adversarial Vulnerability}
Like other DL models, GNNs are inherently vulnerable to adversarial attacks~\cite{sun2024task, sun2023single, fein2023benchmarking}, which hinders their deployment in real-world applications and critical decision-making processes.
\par
Adversarial modifications to graph topology or node features can significantly degrade model performance. For instance, even minor manipulations of pixel values in SAR images can lead to substantial drops in GNN accuracy~\cite{ye2023adversarial}. While there is extensive literature on adversarial attacks in DL for RS scene classification~\cite{xu2020assessing} and object detection~\cite{sun2023threatening}, the adaptation of these concepts to GNNs in the EO domain remains underexplored. Adversarial attacks may involve adding or removing edges, altering node embeddings, or introducing new nodes~\cite{jin2021adversarial, wang2024graph}. Future research should focus on exploring more sophisticated adversarial perturbations and developing corresponding defenses to enhance the robustness of GNNs.

\subsection{Graph Learning Design}
GNN-like models tend to overly smooth predictions with increasing depth~\cite{li2019deepgcns}. In addition, there are diverse and sophisticated GNN models, and each of these methods carries respective strengths. How to select a suitable architecture from a plethora of options requires careful consideration. 
\par
\begin{figure*}[!t]
    \centering
    \includegraphics[width=1.3\columnwidth,trim={0cm 0cm 0cm 4.2cm},clip]{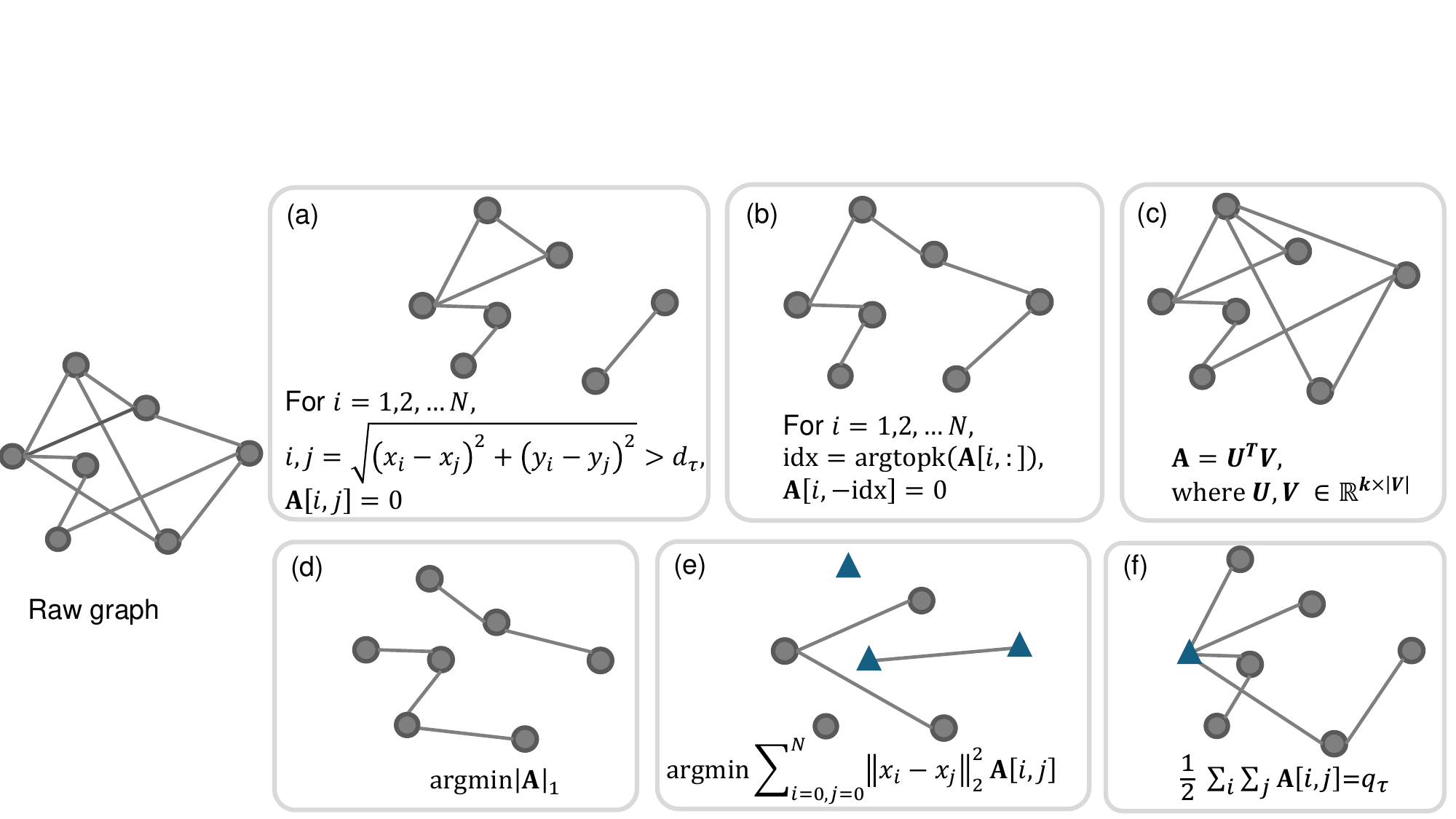}
    \caption{Constraints to train an adjacency matrix. (a) Edge dropping. The edge between distant stations will be dropped according to First Law of Geography (``\textit{Everything is related to everything else, but near things are more related than distant things.}''~\cite{tobler1970computer}). (b) Edge dropping. Only keep K-hop neighbors or top K closet neighbors for each node to filter the most relevant information. (c) Low rank assumption. It is assumed that the intrinstic dimensionality of the data is much lower than the number of nodes in the dataset. The reduce computation complexity is $O(k \times |V| \times d + d^2)$. (d) Sparsity regularization. Minimizing L1 norm to enable the sparsity of the $\mathcal{A}$ matrix. (e) Heterogeneous regularization. The similar nodes are connected with edges (or edges with large weights) and dissimilar nodes without connection (or edges with small weights). (f) Edge dropping. Fix the total number of edges in the graph to be $q_{\tau}$. This allows for sparsity while giving free assignment of node importance. The blue node has higher degree than others, which may be interpreted as more influential location~\cite{cachay2021world}.}
    \label{fig:graph_regularization}
\end{figure*}

To improve the expressiveness of the graph structure, it is common to align it with real-world graphs. Since most real-world graphs are sparse, dropping strategies for more efficient representations are often designed. Besides, regularization techniques for a pursuit of being resemble to graphs with specific properties are introduced during the training procedure. A compendium of these techniques can be found in Fig.~\ref{fig:graph_regularization}.
\par
There are diverse and sophisticated GNN models, each carries respective strengths, and there’s no universally superior approach. Temporal GNNs, for example, can handle the dynamics of EO data, wherein the time-evolving EO data is organized as a sequence of attributed graphs with discrete or continuous time stamps~\cite{rossi2020temporal, bastos2023beyond, han2021dynamic, skarding2021foundations, sahili2023spatio}. Generative GNs target at learning the distributions of given graphs, with methods like GraphMAE~\cite{hou2022graphmae} and S2GAE~\cite{tan2023s2gae} showing significant efficiency in unsupervised learning~\cite{kipf2016variational}. In parallel, contrastive graph learning approaches, such as DGI~\cite{velivckovic2018deep} and GCA~\cite{zhu2021graph}, excel at capturing graph representations by contrasting positive and negative samples, facilitating easy fine-tuning on downstream graphs. Graph U-Net~\cite{gao2019graph} introduces pooling and unpooling operations on graph-structured data, adept at capturing hierarchical features for EO data segmentation.
\par
To sum up, the choice of architecture should be informed by the specific characteristics and requirements of the EO dataset at hand, ensuring an optimal balance between spatial and temporal processing, scalability, and learning efficiency.

\subsection{Computational Complexity}
EO data often comprises a vast number of pixels or elements. This abundance of data, while valuable, presents significant computational challenges. Besides, the nature of message passing in GNNs necessitates communication and synchronization at every step. Last, the dynamic nature of EO data also adds to the complexity. Adding a new node or edge in a graph can have far-reaching effects due to the ``small world'' phenomenon~\cite{watts1998collective} prevalent in many real-world networks. This interconnectedness means that a single change requires a complete recomputation of the predictions or embeddings. The time complexity of training a GCN layer~\cite{kipf2016semi} is related to the sparse-dense matrix multiplication between the adjacency matrix and node features, and normal dense-dense matrix multiplication to the weight matrix~\cite{duan2022comprehensive}.
\begin{equation}
    \mathcal O(L||\mathcal{A}||_0 D + LND^2),
\end{equation}
where $D$ is the averaged node degree and $L$ is the number of layers. Training GNN-based models for large-scale applications demands substantial GPU/TPU resources. For instance, training GraphCast on 30 TB of ERA5 data requires hundreds of TPU training days, as shown in Table~\ref{tab:weather_time}.
\begin{table}[htbp!]
    \caption{\re{The training time for GNN-based, Transformer-based, and hybrid models in global weather forecasting is significant. However, these DL-based models can generate forecasts over 10,000 times faster than traditional physics-based NWP models.}}
    \begin{tabular}{p{1.4cm}  p{1.3cm}  p{1.3cm} p{3.0cm}}
        \hline
        \textbf{Model} & \textbf{Type} & \textbf{Training Time} & \textbf{Inference Time} \\ \hline
        GraphCast~\cite{lam2022graphcast} & GNNs & 28 days $\times$ 32 TPUs & 10-day forecast (at 6-hour steps) takes about 60 seconds on a single Cloud TPU v4 device. \\ \hline
        AIFS~\cite{lang2024aifs} & GNNs with Transformers & 7 days $\times$ 64 A100 GPUs & 10-day forecast takes about 2.5 minutes on a single A100 GPU device. \\ \hline
        GenCast~\cite{price2023gencast} & GNNs with Diffusion models & 5 days $\times$ 32 TPU v5 & A single 15-day forecast takes about 8 minutes on a Cloud TPU v5 device. \\ \hline
        Pangu-Weather~\cite{bi2023accurate} & Transformers & 16 days $\times$ 192 V100 GPUs & A week-long forecast takes less than 2 seconds. \\ \hline
    \end{tabular}
    \label{tab:weather_time}
\end{table}

\par
In other ML domains, some tricks, such as batching or stochastic training methods, are commonly used to reduce computation costs. Nonetheless, their application to geospatial graphs can be non-trivial due to the inherent heterogeneity (non i.i.d.) of the nodes. Removal of certain nodes (subsampling) might lead to a loss of critical information, and the subsampled graph may lack the expressiveness of the global signature. Encoding the geolocations of each node before randomly sampling becomes a solution. Thereafter, stochastic sampling in each training iteration not only supports more robust predictions by preventing the model from memorizing the neighborhood structure, but also allows for computing a ``shuffled'' version of the Moran’s I metric, capturing spatial autocorrelation at arbitrary scales~\cite{klemmer2023positional}.

\subsection{Scalability to Large Graphs}
\re{Deploying GNN in real-world scenarios, such as disaster monitoring and global weather forecasts, necessitates the capacity of real-time processing and large graphs learning. However, the computational complexity and generalizability challenges associated with GNNs often impede their scalability to large-scale applications.}
\par
\re{One primary barrier lies in the computational complexity associated with large-scale graphs. One option to optimize memory and computational efficiency is using sparse matrix operations. Pooling and sampling layers mitigate these challenges, too. Wu \textit{et al.}~\cite{wu2020connecting} subsample nodes from the whole graph. Graph clustering algorithm~\cite{chiang2019cluster} can partition a graph into subgraphs, and by training a GCN on the partitioned sub-graphs, the memory issue is mitigated.}
\par
Besides, the training strategy is also a key component to help generalize to large-scale scenarios. For instance, knowledge distillation has shown to be more effective than traditional training methods in enhancing a model's scalability to large graphs. This approach involves first training a highly expressive GNN as a teacher model, then transferring its knowledge to a simpler student model, such as an multilayer perceptron (MLP)~\cite{zhang2021graph, joshi2022representation}. This process enables easier deployment and faster inference.

%% file: sources/6_discussion.tex
\section{GNNs and Other Neural Architectures}\label{sec:discussion}
While GNNs exhibit considerable representational prowess within non-Euclidean domains, their use in EO applications remains comparatively less widespread. The selection of an appropriate neural network architecture for specific tasks can indeed pose a considerable dilemma. However, it is noteworthy that several prevalent architectures, including CNNs and Transformers, share common characteristics with GNNs when viewed from a high-level perspective. In this regard, we emphasize these shared traits and subsequently delve into the design considerations of respective architectures. Although there is no universally optimal model architecture for EO data, Table~\ref{tab:gnnvsOthers} offers a general comparison by highlighting the key features, advantages, and limitations of each model architecture within the EO context.
Moreover, we explore emerging hybrid techniques in the context of EO applications.

\subsection{GNNs versus CNNs}
\begin{figure}[!t]
    \centering
    \includegraphics[width=1\columnwidth,trim={2cm 1.9cm 8cm 7.5cm},clip]{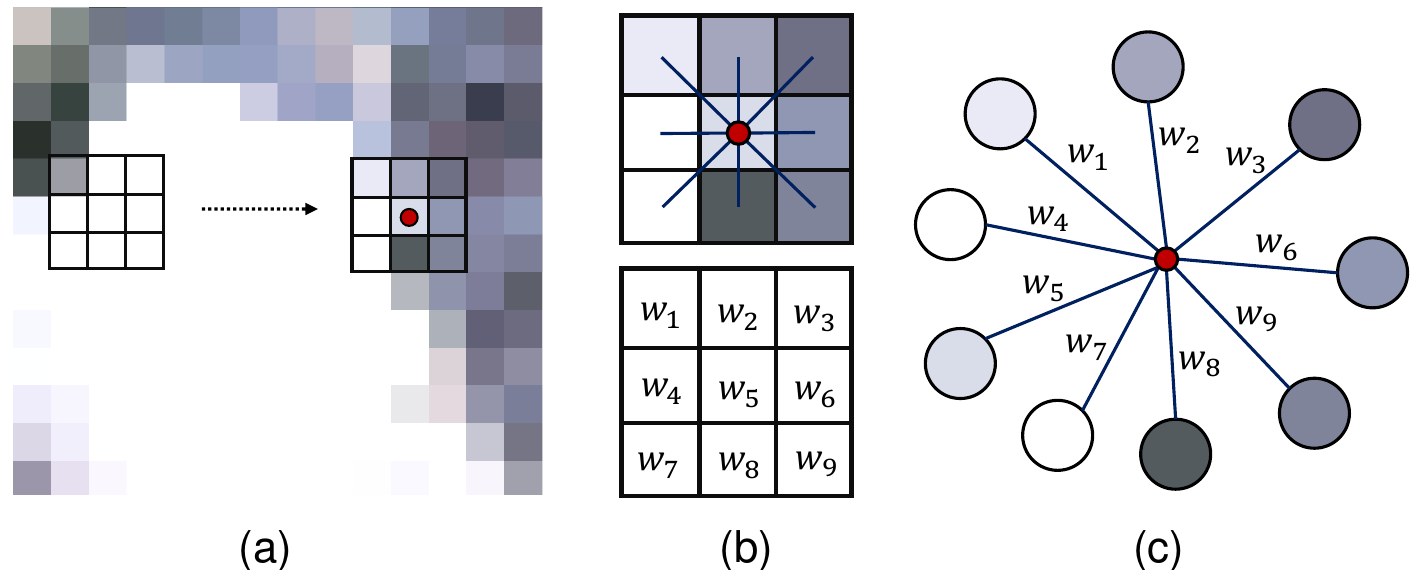}
    \caption{Convolution operation in CNN can be interpreted as graph feature aggregation in GNN. For a convolution kernel applied to an input (a), a graph structure exists where the output at the center connects to neighboring pixels with associated weights (b). This can be formulated as feature aggregation on a graph (c), with edge weights implied by kernel weights.}
    \label{fig:gnn_and_cnn}
\end{figure}

CNNs can be conceptualized as a specialized instance of GNNs as explained in Fig.\ref{fig:gnn_and_cnn}. In this analogy, each pixel in an image serves as a node within a fixed-structure graph, with the convolutional kernel acting as the weighted connections between nodes.
In CNNs, the convolutional operation involves the systematic traversal of the input image using a kernel, where each kernel weight corresponds to a connection weight in the graph. This operation effectively aggregates information from neighboring pixels, akin to the message passing mechanism in GNNs.

However, unlike conventional GNNs where the graph structure is learned from the data, CNNs operate on a predetermined grid-like structure defined by the spatial arrangement of pixels in the image. This fixed structure simplifies the learning process and enables CNNs to efficiently capture spatial dependencies within the image domain.

\subsection{GNNs versus Transformers}
\begin{figure}[!t]
    \centering
    \includegraphics[width=1\columnwidth,trim={1.65cm 0.3cm 3.85cm 0.35cm},clip]{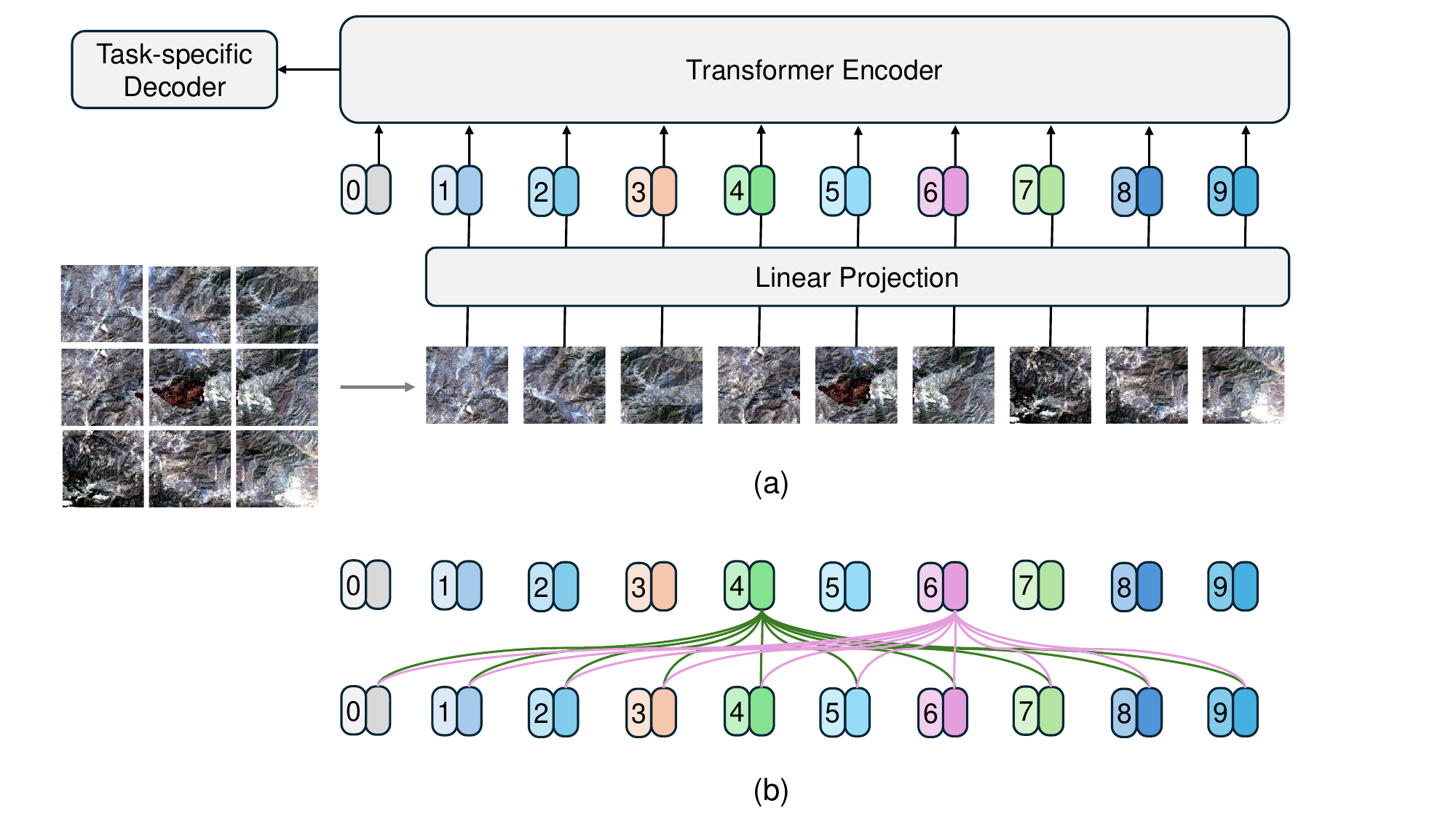}
    \caption{Transformer can be interpreted as a GNN. In Transformers, tokens attend to each other through self-attention mechanisms, where each token aggregates information from all other tokens based on learned attention weights (a). This mechanism is analogous to an exhaustive graph structure (b). For brevity, only two sets of edges are shown in (b).}
    \label{fig:gnn_and_transformer}
\end{figure}
In contrast to GNNs, which explicitly utilize an adjacency matrix to facilitate feature learning, Transformers adopt a distinct approach by implicitly capturing relational information through the attention mechanism, as shown in Fig.\ref{fig:gnn_and_transformer}. The attention mechanism in Transformers extensively models the contributions of pairwise inputs across the entire sequence, thereby enabling the network to selectively attend to relevant context.

The attention mechanism in Transformers operates by computing attention scores between each pair of input elements, typically referred to as keys, queries, and values. These attention scores quantify the importance of each input element with respect to others, allowing the model to focus on relevant information while suppressing irrelevant or noisy signals.

By leveraging the attention mechanism, Transformers effectively capture long-range dependencies and complex interactions within the input sequence, without relying on predefined graph structures. Furthermore, unlike GNNs, which typically operate on fixed-size local neighborhoods defined by the adjacency matrix, Transformers employ self-attention mechanisms that enable each input element to interact with all other elements in the sequence. This holistic approach facilitates information exchange and enables Transformers to effectively model dependencies across the input sequence.
\begin{table*}[htbp]
    \centering
    \caption{\re{The comparison between GNNs, CNNs, RNNs, and Transformers. Each DL architecture offers distinct advantages and limitations when applied to EO data processing.}}
    \label{tab:gnnvsOthers}
    \begin{tabular}{p{2cm} p{3cm} p{6cm} p{4cm}}
    \hline
         Models & Key features & Advantages & Limitations\\\hline
         GNNs & Message passing, label propagation & Irregular data processing, heterogeneous data modeling, contextual information, fine-granularity, semi-supervised learning & Predefine of graph topology, scalability to large graphs \\ \hline
         CNNs & Shift-invariance, locality, parameter sharing & High-efficiency on grided inputs, translation invariance features, deep and local features & Lack of contextual information, limited to fixed grid structures \\ \hline 
         RNNs & Recurrent connections, gates mechanism & Sequential data processing & Limited long-term dependencies and memory, fixed sequential order\\ \hline 
         Transformers & Self-attention mechanism, positional encoding & Long-range interaction, parallel processing & Computationally expensive, requires large training data \\\hline
    \end{tabular}
\end{table*}
\subsection{Hybrid Architectures}
Given that the strengths and weaknesses of different models can complement each other, integrating them into a hybrid architecture holds promise for more effective feature extraction. GNNs have shown superior performance when combined with other architectures such as CNNs, RNNs, and Transformers.
\par
\subsubsection{GNNs with CNNs}
CNNs excel in extracting local features but have a limited receptive field. Conversely, GNNs are adept at capturing long-range contextual information, albeit with a higher computational cost. Combining these models in a hybrid architecture enhances feature extraction at both local and non-local scales, which is particularly relevant in segmentation tasks. For example, in an encoder-decoder architecture, Pei \textit{et al.}~\cite{pei2023application} use a CNN-based encoder for capturing local features and a GCN for exchanging long-range information. Similarly, Dong \textit{et al.}~\cite{dong2022weighted} fuse features from pixel-based CNNs and superpixel-based GATs for multi-scale feature extraction with moderate computational costs. Besides, CNNs can proficiently extract features from inputs as node attributes within a GNN. In Zhang \textit{et al.}~\cite{zhang2022spatiotemporal}, each time series from seismic stations is processed by a CNN, serving as a node feature and thereby incorporating station-specific information into the graph. Fan \textit{et al.}~\cite{fan2022gnn} utilize multiple separate CNNs to process different data types, attaching them as heterogeneous node features. Additionally, CNNs also play the role in building the graph architecture. Zhou \textit{et al.}~\cite{zhou2023deep} employ a CNN to derive the adjacency matrix from the input, creating a learnable architecture that adapts to changing inputs.

\subsubsection{GNNs with RNNs}
Given the dynamic nature of EO data, understanding its temporal patterns holds importance across many applications, such as forecasting and change detection. GNNs are commonly integrated with RNNs for two primary purposes: 1) long and short range information extraction; 2) spatial-temporal pattern capturing.
\par
When it comes to grided data, GNNs offer flexibility in designing graph structures. Therefore, hand-crafting neighboring pixels as connected nodes in GNN promote local information exchange. RNNs, on the other hand, can model long-sequential information. In this regard, the integration of these two components complements each other, preserving both spatial fine-granularity and long-contextual information. Shi \textit{et al.} \cite{shi2020building} conceptualize each pixel represents a node. The adjacency matrix is defined by the spatial neighbours of pixels. The GCN aggregates short-range information from neighbor nodes, allowing the model to learn about local structures. As for long-range dependencies, edge embeddings are taken as hidden states and the updated node representation is used as the input of Gated GRU. With increasing RNN depth, different level features are progressively concatenated. Furthermore, various RNN-based architectures, such as GRU and LSTM, are integrated with GNNs to model both spatial and temporal dynamics. For example, RNN-extracted features can serve as node features within a graph~\cite{zhao2024causal}. Also, GNN-extracted features can be the hidden state of LSTM or GRU modules. Such a combination boosts up their capacity to preserve both spatial and temporal correlations~\cite{michail2024seasonal, xiao2022dual, zanfei2022graph, sit2021short} within EO data.

\subsubsection{GNNs with Transformers}
Considering the synergy between GNNs and Transformers, they can be easily integrated. The integration can be achieved through: \re{1) introducing attention mechanisms to GNNs, 2) incorporating GNNs as part of Transformer blocks or vice versa, and 3) improving the computational efficiency by using graph topology as prior knowledge for attention maps.}
\par
The GAT exemplifies this integration by introducing the attention mechanism into GNNs. When presented with graph-structured data, the masked self-attentional layer automatically computes contributions from different nodes within a neighborhood~\cite{velivckovic2017graph}. The attention mechanism is shared across all nodes, making it scalable for large graph-structured data. Graph Transformer Networks~\cite{yun2019graph} use the attention mechanism to select or generate important connectivities between nodes for a highly representative subgraph generation. Many EO applications have incorporated attention mechanisms into GNN design. For example, Bilal \textit{et al.}~\cite{bilal2022early} use attention mechanisms to focus on important signals from neighboring seismic stations. GRAST-Frost~\cite{lira2021frost} introduces both spatial attention and temporal attention to consider spatial correlation and temporal dynamics simultaneously. Additionally, many attempts have been made to integrate GNNs into transformer architectures. For instance, the message passing layer is instantiated by transformer blocks~\cite{dwivedi2020generalization}. Furthermore, graph structures can serve as a prior to reduce the computational cost of pairwise attention in Transformers, and further efforts are being made towards improving the scalability of Graph Transformers for large graphs~\cite{deng2024polynormer}. \re{AIFS~\cite{lang2024aifs} uses GNN-based encoder and decoder, and modifies the sliding window approach in Swin-Transformer~\cite{liu2021swin} from partitioned 2D squares to slicing on a sphere by latitudes.} This method yields improved fine-grained features compared to a purely GNN-based model~\cite{lam2022graphcast}, and achieves higher computational efficiency than a quadratic memory complexity consumed in traditional Transformers. \re{Table~\ref{tab:weather_time} provides a comparison of the training costs associated with purely GNN-based models, hybrid models, and purely-Transformer based models in the task of global weather forecasting.}

%% file: sources/7_conclusions.tex
\section{Outlooks and Conclusions}
\re{Through a comprehensive overview of the most important works of GNNs in the EO domain, we archive a milestone in the current technical achievements of GNNs within EO field. Despite the substantial success in natural images, much of the potential of GNNs in the EO domain remains untapped. Future research should prioritize integrating the strengths of GNNs with other neural architectures. Furthermore, collaboration between methodological community and Earth Science domains can strengthen cross-disciplinary insights. Such integration presents opportunities to enhance the efficiency, effectiveness, and robustness of GNN models, paving the way for their deployment in real-world applications and driving innovative scientific discoveries. Specifically, we have the following recommendations for future research.}

\subsubsection{Spatial-Temporal Graphs}
Spatial-temporal graphs, despite their complexity, have broader applications in analyzing Earth's dynamic processes. Current methods need to be enhanced to better represent multidimensional data. The modeling of spatial-temporal interactions includes integrating dynamic processes such as river flow patterns, atmospheric transmissions, as well as time-evolving environmental factors.

\subsubsection{Graph Foundation Models}
\re{The vast volume and diversity of EO data offers the opportunity for training graph foundation models. By leveraging the extensive knowledge and resources available during pre-training, GNNs can enhance performance in both supervised and zero-shot graph learning tasks, as exemplified by GraphGPT~\cite{tang2024graphgpt} and OFA~\cite{liu2023one}. This approach is particularly relevant in the EO context, where data labels are sparse and the data spans a wide range of scales (from small to large-scale graphs) and types (including both heterogeneous and homogeneous graphs).}

\subsubsection{Advancing Graphs for Large-Scale Applications}
\re{To promote the deployment of GNNs in real-world applications, it is essential to enhance scalability for large graphs, improve model effectiveness and efficiency. Key steps include collecting representative and realistic EO data, benchmarking graphs for large-scale applications~\cite{colomba2023vigeo}, and developing tailored features for specific use cases. A successful example is global weather forecasting, where progress has been made through architectural innovations (such as integrating GNNs with transformers~\cite{lang2024aifs} or diffusion models~\cite{price2023gencast}), efficiency improvements (like adopting hierarchical GNNs to reduce node numbers~\cite{oskarsson2024probabilistic}), and customized features (e.g., introduce probabilistic models~\cite{oskarsson2024probabilistic, price2023gencast} to GNNs for capturing the uncertainty in the chaotic weather system). The real-time forecast platforms are shown on ECMWF’s OpenCharts\footnotemark \footnotetext{\url{https://charts.ecmwf.int/}} and NOAA/NCEP's Environmental Modeling Center~\cite{wangmachine}. Additional scenarios, such as disaster monitoring and smart city development, could greatly benefit from the operational deployment of GNN models.}

\subsubsection{Discovering New Geoscientific Insights}
\re{Incorporating geoscientific and physical knowledge into GNN models enhances their physical feasibility, leading to more robust and reliable predictions. This approach supports critical decision making, especially under extreme climate change expected in the decades to come. On the other hand, applying new GNN paradigms to EO tasks can encourage advancements in Earth Science. For instance, generative graph learning, which generates novel graphs~\cite{guo2022systematic} from existing distributions, opens avenues for exploring new geoscientific patterns from historical observations.}
\par
We conclude that GNNs hold promise for yielding explicit and transparent insights with a data-driven framework. We envision the respective communities to utilize GNNs to promote the physical alignment of DL methods with the ample knowledge prevalent in Earth Sciences. Together, both can help tackle important challenges in Earth systems, from effective representation learning, robust forecasting, and process understanding, to evaluating and analyzing extreme events.